\def\year{2022}\relax
\documentclass[letterpaper]{article} 
\input{header.tex}
\usepackage{aaai22}  
\usepackage{times}  
\usepackage{helvet}  
\usepackage{courier}  
\usepackage[hyphens]{url}  
\usepackage{graphicx} 
\usepackage{natbib}  
\usepackage{caption} 
\DeclareCaptionStyle{ruled}{labelfont=normalfont,labelsep=colon,strut=off} 
\usepackage{algorithm}
\usepackage{algorithmic}

\usepackage{newfloat}
\usepackage{listings}
\floatstyle{ruled}
\newfloat{listing}{tb}{lst}{}
\floatname{listing}{Listing}
\nocopyright
\title{
\vspace{-10px}
Active Learning of Neural Collision Handler for Complex 3D Mesh Deformations}
\allowdisplaybreaks
\newif\ifreview
\reviewfalse
\ifreview
\author{Anonymous Author(s)}
\affiliations{}
\else
\author{
    Qingyang Tan\textsuperscript{\rm 1},
    Zherong Pan\textsuperscript{\rm 2},
    Breannan Smith\textsuperscript{\rm 3}, 
    Takaaki Shiratori\textsuperscript{\rm 3},
    Dinesh Manocha\textsuperscript{\rm 1}
}
\affiliations{
    \textsuperscript{\rm 1} Department of Computer Science, University of Maryland at College Park \\
    \textsuperscript{\rm 2} Lightspeed \& Quantum Studio, Tencent America \\
    \textsuperscript{\rm 3} Facebook Reality Labs Research\\
    qytan@umd.com, zherong.pan.usa@gmail.com, breannan@fb.com, tshiratori@fb.com, dmanocha@umd.edu
}
\fi

\begin{document}

\maketitle

\begin{abstract}
We present a robust learning algorithm to detect and handle collisions in 3D deforming meshes. Our collision detector is represented as a bilevel deep autoencoder with an attention mechanism that identifies colliding mesh sub-parts. We use a numerical optimization algorithm to resolve penetrations guided by the network. Our learned collision handler can resolve collisions for unseen, high-dimensional meshes with thousands of vertices. To obtain stable network performance in such large and unseen spaces, we progressively insert new collision data based on the errors in network inferences. We  automatically label these data using an analytical collision detector and progressively fine-tune our detection networks. We evaluate our method for collision handling of complex, 3D meshes coming from several datasets with  different shapes and topologies, including datasets corresponding to dressed and undressed human poses, cloth simulations, and human hand poses acquired using multiview capture systems. Our approach outperforms supervised learning methods and achieves $93.8-98.1\%$ accuracy compared to the groundtruth by analytic methods. Compared to prior learning methods, our approach results in a $5.16\%-25.50\%$ lower false negative rate in terms of collision checking and a $9.65\%-58.91\%$ higher success rate in collision handling.
\end{abstract}

\begin{figure*}[t]
\vspace{-20px}
\centering
\includegraphics[width=0.7\textwidth]{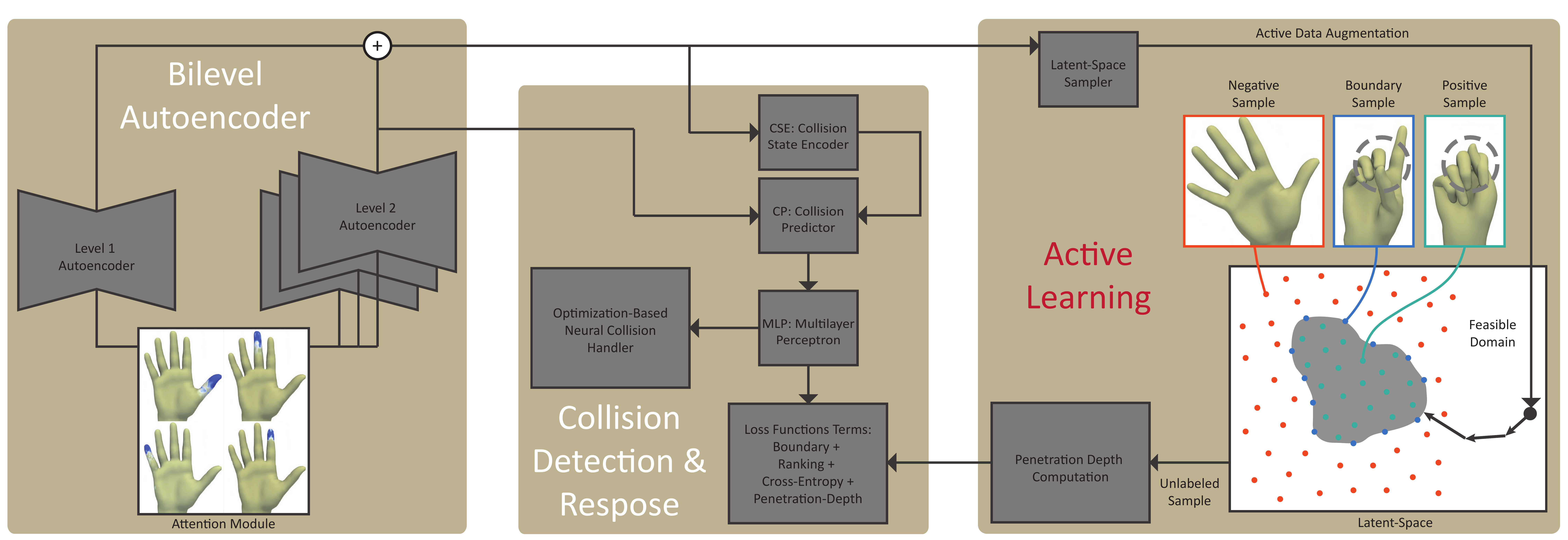}
\vspace{-10px}
\caption{\footnotesize{\label{fig:pipeline} We highlight the three main components of our approach. From left to right: the bilevel autoencoder \cite{yang2020multiscale}, the neural collision detector and optimization-based collision handler \cite{tan2020lcollision}, and active learning method (our contribution). Our method progressively insert data by randomly sampling in the latent space (top right). We then use a Newton-type method to pull samples towards the learned decision boundary (black arrows in the white, feasible domain). This risk-seeking method can best help our neural collision detector improve its accuracy. The groundtruth collision labels are generated using an analytic collision detector (bottom right). Finally, we use three different loss functions for samples on the positive (orange), negative side (green), and near (blue) the decision boundaries. These three techniques combined make our neural collision detector much more robust than state-of-the-art \cite{tan2020lcollision}.}}
\vspace{-20px}
\end{figure*}

\section{\label{sec:intro}Introduction}
Learning to model or simulate deformable meshes is becoming an important topic in computer vision and computer graphics, with rich applications in real-time physics simulation \citep{holden2019subspace}, animation synthesis \citep{9217964}, and cross-domain model transformation \citep{VOCA2019}. Central to these methods are generative models that map high-dimensional deformed 3D meshes with rich details into low-dimensional latent spaces. These generative models can be trained from high-quality groundtruth datasets, and they infer visually or physically plausible meshes in real time. These 3D datasets can also be generated using physics simulations \citep{Narain:2012:AAR,tang2012continuous} or reconstructed from the physical world using multi-view capture systems \citep{10.1145/3414685.3417768}. In general, 3D deformable meshes are more costly to acquire, so 3D mesh datasets typically come in smaller sizes than image or text datasets. Inference models trained using such small datasets can suffer from over-fitting and generate meshes with various visual artifacts. For example, human pose embedding networks \citep{tan2018mesh,gao2018automatic} can have excessive deformations, and interaction networks \citep{battaglia2016interaction} can result in non-physically-based object motions. 

Our main goal is to resolve a major source of visual artifacts: self-collisions. Instead of acquiring more data, we argue that domain-specific knowledge could also be utilized to significantly improve the accuracy of inference models. There have been several prior research works along this line. For example, \citep{DBLP:journals/corr/abs-2008-05440} exploited the fact that near articulated meshes can be divided into multiple components, and they train a recursive autoencoder to stitch the components together. \citep{Zheng_2021_CVPR} utilized the locality of secondary physics motions to learn re-targetable and scalable real-time dynamics animation. Recently, \citep{tan2020lcollision} studied learning-based collision avoidance for 3D meshes corresponding to human poses. They proposed a deep architecture to detect collisions and used numerical optimizations to resolve detected collisions. However, \citep{tan2020lcollision} used a large mesh dataset to obtain stable performance of neural collision detection. Indeed, a deformed 3D mesh typically involves more than $10^4$ elements (voxels, points, triangles) where any pair of two elements can have collisions. Therefore, a huge amount of data is required to present the inference model with enough examples of collisions between all possible element pairs.

\TE{Main Results:} We present a robust method to train neural collision handler for complex, 3D deformable meshes using active learning. Our key observation is that the distribution of penetrating meshes can have a long tail and active learning is an effective method for modeling the tail \cite{geifman2017deep}. Specifically, most penetrating mesh primitives have large overlapping patches consisting of the central part of the distribution, but many other meshes have small patches that penetrate each other, forming the tail. In fact, most 3D mesh datasets do not focus on generating samples in the tail and cannot be used to train stable collision detectors. In order to overcome these issues, our approach combines three main ideas: 1) We use active learning to progressively insert new samples into the dataset. With the help of exact collision detectors, collision labels for the new samples are automatically generated; 2) We use a risk-seeking approach to prioritize samples in the tail, so that the inserted samples can best help the neural collision detector improve its accuracy; 3) We use different loss functions for samples far from and close to the decision boundary. Our overall approach is shown in \prettyref{fig:pipeline}. We apply our training method on the same neural collision detector as \citep{tan2020lcollision}, and we compare with their supervised learning approach on five deformable mesh datasets corresponding to dressed and undressed human poses, cloth simulations, and human hand poses. By comparison, our approach exhibits much higher data efficacy, shows higher accuracy (up to $98.1\%$) in terms of collision detection, and uses much fewer training samples (up to  $48.12\%$ fewer). Given a dataset of the same size, our neural collision pipeline reduces the false negative rate by $14.27\%$ on average, and we successfully resolve $24.74\%$ more self-colliding meshes. Overall, ours is the first practical method for neural collision handling for complex 3D meshes.
\section{\label{sec:related}Related Work}
Our method is designed for general deformable meshes with fixed topology. We first learn a latent-space of meaningful mesh deformations and then use active learning to train a neural collision detector that identifies a collision-free subspace. We review related work in these areas.

\TE{Generative Model of Dense 3D Shapes:} Categorized by shape representations, generative models can be based on point clouds \citep{qi2017pointnet++}, volumetric grids \citep{wu20153d}, multi-charts \cite{groueix2018papier}, surface meshes \citep{tan2018variational}, or semantic data structures \cite{10.1145/3306346.3322969}. We use mesh-based representations with fixed topologies, because most collision detection libraries are designed for meshes. Note that this choice excludes several applications that require general meshes of changing topology, e.g., for modeling meshes of hierarchical structure \citep{yu2019partnet} or modeling scenes with many objects \citep{ritchie2019fast}. However, these applications typically involve only static meshes with no need for collision detection. There is a separate research direction on domain-specific mesh deformation representation, e.g., SMPL/STAR human models \citep{SMPL:2015,STAR:2020}, wrinkle-enhanced cloth meshes \citep{lahner2018deepwrinkles}, and skeletal skinning meshes \citep{RigNet}. There are even prior works \citep{Fieraru_2020_CVPR,fieraru2021learning,Muller_2021_CVPR} on collision detection and handling for human bodies. These domain-specific methods are typically more accurate than our representation, but by assuming general meshes, our representation can be applied to multiple domains as shown in \prettyref{sec:evaluation}.

\TE{Collision Prediction \& Handling:} Although collision detection has been well studied and mature software packages are available, detecting and handling self-collisions can still be a non-trivial computational burden for large meshes. Prior methods \citep{pan2012fcl,kim2018chapter,govindaraju2005quick} use spatial hashing, bounding volume hierarchies, and GPU to accelerate the computation by pruning non-colliding primitives, but they cannot be generalized to learning methods. Handling collisions is even more challenging, and prior physically-based methods either use penalty forces coupled with discrete collision detectors \citep{tang2012continuous} or hard constraints coupled with continuous collision detectors \citep{Narain:2012:AAR}. All these methods rely on physics-based constraints to handle collisions. Recently, many learning methods such as \citep{Gundogdu_2019_ICCV,patel20tailornet} have been designed to predict the cloth movement or deformation in 3D, but they do not perform collision handling explicitly.

\TE{Active Learning:} An active learner alternates between drawing new samples and exploiting existing samples. These samples can be drawn guided by an acquisition function in Bayesian optimization \citep{niculescu2006bayesian} or from an expert algorithm \citep{de2018learning}. Active learning has been applied to approximate the boundary of the configuration space \citep{pan2013efficient,tian2016efficient,das2017fastron}, where the feasible domain of collision constraints is parameterized using kernel SVM. However, this method is limited to rigid or articulated deformation and is not applicable to general 3D deformations. More broadly, active learning has been adopted in various prior works to accelerate data labeling in image classification \citep{gal2017deep} and object detection \citep{aghdam2019active} tasks. These methods progressively identify unlabeled images to be forwarded to experts for labelling. An alternative method for selecting the samples is identifying a coreset \citep{paul2014visual}, and authors of \citep{sener2018active} propose a practical algorithm for coreset identification via k-center clustering. While these methods consider a pre-existing set of unlabeled images for training discriminative models, we assume a continuous latent space of samples for training generative models and use a risk-seeking method to identify critical new samples. Further, we automatically label the new samples using analytical collision detection algorithms, making the training method fully automatic.
\begin{table}
\vspace{-5px}
\resizebox{0.47\textwidth}{!}{
\begin{tabular}{cc}
\rowcolors{0}{gray!50}{white}
\begin{tabular}{ll}
\toprule
Variable & Definition \\
\midrule
$\mathcal{G}=\left<V,E\right>$ & graph with vertices and edges\\
$E$, $D$ & encoder, decoder\\
$\theta_{E,D,C}$ & learnable parameters\\
$Z_\text{all}=\FOUR{Z_0}{Z_1}{\cdots}{Z_{|Z_0|}}$ & autoencoder latent code\\
$\mathcal{Z}$ & latent region of sampling\\
PD & penetration depth\\
CSE & global collision state encoder\\
CP & local collision state predictor\\
$\text{MLP}_c$ & collision classifier\\
$S_0, S_1, \cdots, S_{|Z_0|}$ & neural collision indicator\\
\hline
\end{tabular}
&
\rowcolors{0}{gray!50}{white}
\begin{tabular}{ll}
\toprule
Variable & Definition \\
\midrule
$\text{ACAP}$ & feature transform function\\
$\theta_C$ & neural collision network parameters\\
$I_c$ & collision state label\\
$E$ & collision handler objective function\\
$\mathcal{D}$ & dataset for learning $\theta_{E,D}$\\
$\mathcal{D}_{d,p,n,c}$ & dataset for learning $\theta_C$\\
$\epsilon$ & threshold for boundary samples\\
CE & cross-entropy loss\\
$\mathcal{L}_\bullet$& neural network losses\\
$w_\bullet$ & weights for each loss\\
\hline
\end{tabular}
\end{tabular}
}
\vspace{-10px}
\end{table}
\section{\label{sec:problem}Neural Collision Handler}
In this section, we briefly review the mesh-based generative models with a neural collision handler \citep{tan2020lcollision}, based on which we build our active learning method. All notations are summarized in the symbol table. 

A mesh is represented by the graph $\mathcal{G}=\left<V,E\right>$, where $V$ is a set of vertices, and $E$ is a set of edges. We assume that all the meshes have the same topology, that is, all the meshes differ in $V$ while the connectivity $E$ stays the same. We further limit ourselves to manifold triangle meshes, i.e., each edge is incident to at most two triangles, and two triangles are adjacent if and only if they share an edge. No other assumptions are made on the mesh deformation. We denote a mesh as self-collision-free if and only if any pair of two non-adjacent triangles are not intersecting each other. Our goal is to design a mesh-based generative neural architecture where we take as input a coordinate in the latent space and output a 3D mesh without self-collisions. The latent space is defined as a low-dimensional space that can be mapped to high-dimensional meshes injectively using a learned decoder function (\prettyref{fig:pipeline} right). Furthermore, the latent-to-mesh mapping is differentiable and supports multiple downstream applications explained in \prettyref{sec:app}. Our method consists of two parts. First, we train a bilevel mesh autoencoder using supervised learning. Second, we train a neural collision handler using active learning and a special boundary loss.

\subsection{Bilevel Autoencoder}
Our bilevel autoencoder architecture maps a deformed mesh to two levels of latent codes. The latent codes are used to both recover the high-dimensional deformed mesh vertices $V$ and predict whether the deformed mesh contains self-collisions. We only want to encode intrinsic mesh information such as curvatures instead of extrinsic rigid transformations because mesh shapes are invariant to extrinsic transformation. Therefore, we first use as-consistent-as-possible (ACAP) feature transformation \cite{gao2019sparse} to factor out rigid transformations. The ACAP feature vector is first brought through the level-1 autoencoder and mapped to a latent code $Z_0$. Since our dataset size is small, we use a shallow autoencoder to avoid over-fitting and $Z_0$ is subject to large embedding error. We further hypothesize that the error is sparsely distributed throughout the mesh vertices. Therefore, we use an attention mechanism trained with a sparsity prior to decompose the mesh into near-rigid sub-domains. The sparsity prior is designed such that each domain can be mapped to a single axis of the latent space, i.e., a single entry of $Z_0$. Afterwards, a set of $|Z_0|$ level-2 autoencoders is introduced to further reduce the error (\prettyref{fig:pipeline} left), with each autoencoder dedicated to one entry of $Z_0$. Their latent codes are denoted as $Z_1,\cdots,Z_{|Z_0|}$. The ultimate mesh is reconstructed from $Z_\text{all}$ by combining level-1 and level-2 latent codes:
\small
\begin{align*}
Z_\text{all}=&
\FOUR{Z_0}{Z_1}{\cdots}{Z_{|Z_0|}}\\
D(Z_\text{all},\theta_D)=&
\sum_{i=0}^{|Z_0|}D_i(Z_i,\theta_{D_i})\\
V=&\text{ACAP}^{-1}(D(Z_\text{all},\theta_D)),
\end{align*}
\normalsize
where $D_i$ is the $i$th decoder, with $\theta_{D_i}$ being the learnable parameters. The mesh vertices $V$ are reconstructed by inverting the ACAP transformation. Correspondingly, we have the encoder defined as $E(\text{ACAP}(V),\theta_E)=Z_\text{all}$, which maps the vertices of a mesh to the latent space, with $\theta_E$ being the learnable parameters. 

\subsection{Neural Collision Detector}
\setlength{\columnsep}{8pt}
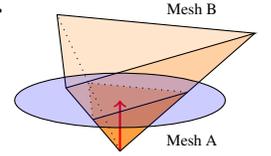
\begin{wrapfigure}{r}{0.19\textwidth}
\centering
\vspace{-35px}
\tdplotsetmaincoords{75}{110}
\begin{tikzpicture}[tdplot_main_coords,scale=1.4]
\coordinate (a) at (  0, 0,-0.5);
\coordinate (b) at (1.5, 0, 0.5);
\coordinate (c) at ( -1, 1, 0.5);
\coordinate (d) at ( -1,-1, 0.5);
\coordinate (b0) at ($(a)!0.5!(b)$);
\coordinate (c0) at ($(a)!0.5!(c)$);
\coordinate (d0) at ($(a)!0.5!(d)$);
\fill[fill=orange,fill opacity=0.2,->] (b) -- (c) -- (c0) -- (b0) -- cycle;
\fill[fill=orange,fill opacity=0.2,->] (c) -- (d) -- (d0) -- (c0) -- cycle;
\fill[fill=orange,fill opacity=0.2,->] (b) -- (d) -- (d0) -- (b0) -- cycle;
\fill[fill=orange,fill opacity=0.5,->] (a) -- (c0) -- (b0) -- cycle;
\fill[fill=orange,fill opacity=0.5,->] (a) -- (d0) -- (c0) -- cycle;
\fill[fill=orange,fill opacity=0.5,->] (a) -- (d0) -- (b0) -- cycle;
\draw[color=red,thick,->] (a) node {} -- (0,0,0) node {};
\draw[] (a) -- (b); 
\draw[] (a) -- (c); 
\draw[dotted] (a) -- (d); 
\draw[] (b) -- (c) -- (d) -- cycle;
\draw[] (b0) -- (c0);
\draw[dotted] (c0) -- (d0);
\draw[dotted] (d0) -- (b0);
\fill[fill=blue,fill opacity=0.2,->] (0,0) circle[radius=1];
\draw[->] (0,0) circle[radius=1];
\node[] at (-2,0,-0.9) {\tiny Mesh A};
\node[] at (-2,0, 0.4) {\tiny Mesh B};
\end{tikzpicture}
\vspace{-15px}
\caption{\footnotesize{\label{fig:PD} PD (red arrow) is the locally minimal translation for mesh B (orange) to be collision-free from mesh A.}}
\vspace{-5px}
\end{wrapfigure}
Our neural collision detector predicts whether the mesh $V$ is subject to self-collisions using latent information $Z_\text{all}$. The extent to which two meshes collide can be measured by the notion of Penetration Depth (PD)~\cite{zhang2014continuous}, defined by the norm of the smallest configuration change needed for a mesh to be self-collision-free, as illustrated in \prettyref{fig:PD}. It is well-known that PD is a non-smooth function of $V$ (esp. at the boundaries), thereby making it difficult to resolve collisions by minimizing PD. By choosing appropriate activation functions ($\tanh$ and CELU in our case), we design the neural collision detector to be a differentiable approximation of PD. As a result, gradient information can be propagated to a collision handler to minimize PD. 

Since collisions can happen between any pairs of geometric mesh primitives, a collision detector should consider possible contacts between any pair of near-rigid sub-domains, leading to a quadratic complexity $\mathcal{O}(|Z_0|^2)$. We use a global-local detection architecture that effectively reduces the number of learnable parameters. Specifically, we introduce a global collision state encoder $S_0=\text{CSE}(Z_\text{all},\theta_C)$ and a set of local collision predictors $S_i=\text{CP}(S_0,Z_i,\theta_C)$, with $i=1,\cdots,|Z_0|$, which predicts whether the $i$th sub-domain is in collision with the rest of the mesh. Finally, the collision information for all local collision predictors is summarized using a classifier network $\text{MLP}_c(S_1,\cdots,S_{|Z_0|})$ to derive a single overall collision classifier:
\small
\begin{align*}
S_0\triangleq&\text{CSE}(Z_\text{all},\theta_C)\\
S_i\triangleq&\text{CP}(S_0,Z_i,\theta_C)\\
I_c(Z_\text{all})\triangleq&\mathbb{I}(\text{MLP}_c(S_1,\cdots,S_{|Z_0|},\theta_C)\geq0.5),
\end{align*}
\normalsize
where $\theta_C$ is the learnable parameters. The feasible space boundary of collision-free constraints corresponds to the $0.5$-levelset of $\text{MLP}_c$. This architecture combining $\text{CSE}$, $\text{CP}$, and $\text{MLP}$ has several learnable parameters proportional to $\mathcal{O}(|Z_0|)$.

\subsection{Optimization-Based Collision Response\label{sec:app}}
Existing collision handling techniques \cite{tang2012continuous,Narain:2012:AAR} are mostly based on numerical optimizations, where a key challenge is to make sure that collision constraints are differentiable. \citep{tan2020lcollision} also uses this formulation, but it is guided by the learned collision detector $\text{MLP}_c$, which is differentiable by construction. Suppose we take as input a randomly sampled latent code $Z_\text{all}^\text{user}$, which might not satisfy the collision-free constraints. We then need to project that latent code back to the feasible domain of collision-free meshes. We achieve this by solving the following optimization problem under neural collision-free constraints using the Augmented Lagrangian Method (ALM):
\begin{align}
\label{eq:handler}
\argmin{Z_\text{all}}\;E(Z_\text{all})\quad\ST\;
\text{MLP}_c(S_1,\cdots,S_{|Z_0|},\theta_C)\leq0.5,
\end{align}
where $E(\bullet)$ is some objective function, which can take multiple forms, as specified by downstream applications. In the simplest case, we take as input a desired $Z_\text{all}^\text{user}$, and we can define $E(Z_\text{all})=\|Z_\text{all}-Z_\text{all}^\text{user}\|^2/2$, which is only related to latent space variables. As a more intuitive interface, the user might want to change meshes in the Cartesian space instead of the of latent space. For example, if the user wants a human hand to be at a certain position $V^\text{user}$, we could define $E(Z_\text{all})=\|\text{ACAP}^{-1}(D(Z_\text{all},\theta_D))-V^\text{user}\|^2/2$. A desirable feature of \prettyref{eq:handler} is an invariant problem size. However many vertices a mesh has, there is only one constraint, which guarantees high test-time performance. Moreover, it has been shown in \cite[Theorem~10.4.3]{sun2006optimization} that ALM either finds a feasible solution or returns an infeasible solution that is closest to the boundary of the feasible domain. In other words, ALM always makes a best effort to resolve collisions, even if feasible solutions are not available.
\section{\label{sec:method}Active Learning Algorithm}
The goal of active learning is to iteratively improve the accuracy of the neural collision detector. We assume the availability of an existing dataset $\mathcal{D}$ of ``high-quality'' meshes with deformed vertices $\mathcal{D}=\{V^{1,2,\cdots,|\mathcal{D}|}\}$, which is used to train a pair of autoencoders ($\left<\theta_E,\theta_D\right>$) via reconstruction loss. We further assume all the groundtruth meshes are (nearly) self-collision free. However, the autoencoders can still suffer from remaining embedding error after training, and users might explore the latent space in regions that are not well covered by the training dataset. All these factors can lead to self-collisions, which should be recognized by our neural collision detector. Therefore, our network cannot be trained with $\mathcal{D}$ alone. This is because $\mathcal{D}$ only contains negative (collision-free) samples, while the neural collision detector must learn the decision boundary between positive and negative samples. In other words, the network must be presented with enough samples to cover all possible latent codes with both self-penetrating and collision-free meshes. We denote the training dataset of neural collision detectors as another set: $\mathcal{D}_c=\{\left<Z_\text{all}^i,I_c^*(Z_\text{all}^i)\right>|i=1,2,\cdots,|\mathcal{D}_c|\}$, where $I_c^*(\bullet)$ is the groundtruth $0-1$ collision state label.

It has been shown in \cite{gal2017deep,aghdam2019active} that many data points in a large image dataset are similar and that human labeling for each point is time-consuming and contains redundant work. In our case, the groundtruth collision state label can be generated automatically using a robust algorithm such as \cite{pan2012fcl}, to compute $\text{PD}$,  where a positive $\text{PD}$ indicates self-collisions, so we can define $I_c^*(Z_\text{all})\triangleq\mathbb{I}(\text{PD}(\text{ACAP}^{-1}(Z_\text{all},\theta_D))>0)$. However, the cost to compute penetration depth, $\text{PD}(\bullet)$, is superlinear in the number of mesh vertices, and computing $\text{PD}$ for an entire dataset can still be a computational bottleneck. Moreover, we are considering a continuous space of possible training data that cannot be enumerated. To alleviate the computational burden, we design a three-stage method, as illustrated in \prettyref{fig:pipeline}. During the first stage of bootstrap, we sample an initial boundary set by which we train $\text{MLP}_c$ to approximate the true decision boundary. At the second stage of data augmentation, new training data is selected and progressively injected into a dataset. Finally, for the third stage, our neural collision detector is updated to fit the augmented dataset. The criterion for selecting the subset is critical to the performance of active learning. We observe that our neural collision predictor is used as constraints for nonlinear optimization methods so that samples far from the boundary are not used by the optimizer and only the boundary of the feasible domain (gray area in \prettyref{fig:pipeline} right) is useful. Therefore, we propose using a Newton-type risk-seeking method to push the samples towards the decision boundary. We provide more details for each step below.

\subsection{\label{sec:bd}Bootstrap}
Active learning would progressively populate $\mathcal{D}_c$, so prior work \cite{aghdam2019active} simply initializes the dataset to an empty set. However, we find that a good initial guess can significantly improve the convergence of training. This is because we select new data by moving (randomly sampled) latent codes towards the decision boundary of the $\text{PD}$ function using a risk-seeking method. However, the true boundary of the collision-free constraints corresponds to the boundary of $C$-Obstacles, which is high-dimensional and unknown to us ($\text{PD}$ is a non-smooth function, so we cannot even use gradient information to project a mesh to the zero level-set of $\text{PD}$). Instead, we propose using the learned neural decision boundary, i.e., the $0.5$-levelset of $\text{MLP}_c$, as an approximation. If we initialize $\mathcal{D}_c=\emptyset$, the surrogate decision boundary is undefined, and the training might diverge or suffer from slow convergence. For our bootstrap training, we uniformly sample a small set of $N_\text{init}$ latent codes $\mathcal{Z}_\text{all}$ at random positions from the latent space and compute $\text{PD}$ for each of them. We define a valid space of sampling by mapping all the data $Z_\text{all}^i\in\mathcal{D}$ to their latent codes and compute a bounded box in the latent space:
\small
\begin{align*}
\mathcal{Z}=\prod_{j=0}^{|Z_0|}
\left[\fmin{i=1,\cdots,|\mathcal{D}|}\left[e_j^TZ_\text{all}^i\right],
\fmax{i=1,\cdots,|\mathcal{D}|}\left[e_j^TZ_\text{all}^i\right]\right].
\end{align*}
\normalsize
We hypothesize that all the meshes can be embedded using our autoencoder with small error corresponding to latent codes in $\mathcal{Z}$, so we can initialize $\mathcal{D}_c=\{Z_\text{all}^{1,\cdots,N_\text{init}}|Z_\text{all}^i\sim U(\mathcal{Z})\}$. We then divide the data points into three subsets ($\mathcal{D}_c=\mathcal{D}_p\bigcup\mathcal{D}_n\bigcup\mathcal{D}_b$, illustrated in \prettyref{fig:pipeline} right):
\small
\begin{equation}
\begin{aligned}
\label{eq:subset}
\mathcal{D}_p\triangleq&\{\left<Z_\text{all}^i,I_c(Z_\text{all}^i)\right>|
\text{PD}(\text{ACAP}^{-1}(Z_\text{all},\theta_D))>\epsilon\}\\
\mathcal{D}_n\triangleq&\{\left<Z_\text{all}^i,I_c(Z_\text{all}^i)\right>|
\text{PD}(\text{ACAP}^{-1}(Z_\text{all},\theta_D))<0\}\\
\mathcal{D}_b\triangleq&\{\left<Z_\text{all}^i,I_c(Z_\text{all}^i)\right>|
\text{PD}(\text{ACAP}^{-1}(Z_\text{all},\theta_D))\in[0,\epsilon]\}.
\end{aligned}
\end{equation}
\normalsize
Here, $\mathcal{D}_p$ is the positive set consisting of samples with penetrations deeper than a threshold $\epsilon$, $\mathcal{D}_n$ is the negative set consisting of collision-free samples, and $\mathcal{D}_b$ is a boundary set where samples are nearly collision-free and lie on the decision boundary. We propose using the $l_1$-loss function for $\mathcal{D}_b$ to approximate the decision boundary:
\begin{align}
\mathcal{L}_b=E_{\{Z_\text{all}\in\mathcal{D}_b\}}
\left[\|\text{MLP}_c(S_1,\cdots,S_{|Z_0|},\theta_C)-0.5\|\right].
\label{eq:bd_loss}
\end{align}

\subsection{Data Aggregation}
The accuracy of our neural collision detector can be measured by the discrepancy between the surrogate decision boundary deemed by $\text{MLP}_c$ and the true decision boundary of $\text{PD}$, formulated as:
\begin{align*}
E_{\{Z_\text{all}\sim\mathcal{Z}|\text{PD}(\text{ACAP}^{-1}(Z_\text{all},\theta_D))=0\}}
\left[\text{CE}(I_c(Z_\text{all}),I_c^*(Z_\text{all}))\right],
\end{align*}
which is an expectation over the true decision boundary. Here $\text{CE}$ is the cross-entropy loss. However, it is very difficult to derive a sampled approximation of the above metric because $\text{PD}$ is a non-smooth function whose level-set is measure-zero, which corresponds to the boundaries of C-obstacles. Instead, we propose to take expectation over the surrogate decision boundary:
\begin{align*}
E_{\{Z_\text{all}\sim\mathcal{Z}|\text{MLP}_c(S_1,\cdots,S_{|Z_0|},\theta_C)=0.5\}}
\left[\text{CE}(I_c(Z_\text{all}),I_c^*(Z_\text{all}))\right].
\end{align*}
Generally speaking, the $0.5$-level-set of $\text{MLP}_c$ can also be measure-zero, but we have designed our neural networks $D,\text{CSE},\text{CP},\text{MLP}_c$ to be differentiable functions. As a result, we could always project samples onto the $0.5$-level-set by solving the following risk-seeking unconstrained optimization: 
\begin{align*}
\argmin{Z_\text{all}}\;\frac{1}{2}\|\text{MLP}_c(S_1,\cdots,S_{|Z_0|},\theta_C)-0.5\|^2.
\end{align*}
We adopt the quasi-Newton method and update $Z_\text{all}$ using the following recursion:
\footnotesize
\begin{align}
\label{eq:newton}
Z_\text{all}-\bar{H}^{-1}\nabla\text{MLP}_c(\text{MLP}_c-0.5),
\end{align}
\normalsize
where $\bar{H}$ is some first-order approximation of the Hessian matrix, which is much faster to compute than the exact Hessian, which requires the second-order term $\nabla^2\text{MLP}_c$. In summary, we would sample a new set of size $N_\text{aug}/2$ from previous $\mathcal{D}_c$ during each iteration of data augmentation. For each sampled $Z_\text{all}$, we project $Z_\text{all}$ to the surrogate decision boundary using recursive \prettyref{eq:newton} until the relative change within $Z_\text{all}$ is smaller than $\epsilon_z$ between consecutive iterations. We also sample $N_\text{aug}/2$ directly from $U(\mathcal{Z})$, using random samples to discover uncovered regions, which achieves a balance between exploitation and exploration. Finally, we classify $Z_\text{all}$ into either one of $\mathcal{D}_{p,n,b}$, according to \prettyref{eq:subset} using the penetration depth.

\subsection{Model Update}
After $\mathcal{D}_c$ has been updated, we fine-tune $\text{CSE},\text{CP},\text{MLP}_c$ by updating $\theta_C$ using the following loss functions:
\begin{align*}
\mathcal{L}=w_\text{PD}\mathcal{L}_\text{PD}+w_r\mathcal{L}_r+w_\text{ce}\mathcal{L}_\text{ce}+w_b\mathcal{L}_b,
\end{align*}
where $w_\bullet$ are weights corresponding to each type of loss. Our first term $\mathcal{L}_\text{PD}$ is a regularization that enforces consistency between $S_i$ and true $\text{PD}$, defined as:
\small
\begin{align*}
\mathcal{L}_\text{PD}=
E_{\{Z_\text{all}\in\mathcal{D}_C\}}
\left[\sum_{i=1}^{|Z_0|}\|S_i-\text{PD}_i\|^2+w_\text{PDsum}\|\sum_{i=1}^{|Z_0|}S_i-\text{PD}\|^2\right],
\end{align*}
\normalsize
where we penalize both domain-decomposed penetration depth $\text{PD}_i$ defined in \cite{tan2020lcollision} and total penetration depth with weight $w_\text{PDsum}$. Our second term $\mathcal{L}_r$ is a marginal ranking loss that enforces the correct ordering of penetration depth to avoid over-fitting, which is defined as:
\small
\begin{align*}
\mathcal{L}_r=E_{\{Z_\text{all}^{a,b}\in\mathcal{D}_C|\text{PD}^a<\text{PD}^b\}}
\left[\max(0,\alpha-(\sum_{i=1}^{|Z_0|}S_i^a-\sum_{i=1}^{|Z_0|}S_i^b)\right],
\end{align*}
\normalsize
where $\alpha$ is the maximal allowable order violation. We use superscripts to distinguish two samples drawn from $\mathcal{D}_C$. Our third term measures the discrepancy between $\text{MLP}_c$ and $\text{PD}$ over the entire latent space:
\small
\begin{align*}
\mathcal{L}_\text{ce}=
E_{\{Z_\text{all}\in\mathcal{D}_p\bigcup\mathcal{D}_n\}}
\left[\text{CE}(I_c(Z_\text{all}),I_c^*(Z_\text{all}))\right].
\end{align*}
\normalsize
We update our neural collision detector with objective function $\mathcal{L}$ by running a fixed number of training epochs, denoted as $N_\text{epoch}$, with $\theta_C$ warm-started from the last iteration of the model update. Our overall training method is illustrated in \prettyref{alg:activeLearning}.
\begin{algorithm}
\caption{Learning $\theta_E,\theta_D,\theta_C$}
\label{alg:activeLearning}
\begin{algorithmic}[1]
\begin{footnotesize}

\STATE Prepare initial data set $\mathcal{D}$
\STATE Update $\theta_E,\theta_D$ using reconstruction loss and $\mathcal{D}$
\STATE Initialize $\mathcal{D}_C$ by drawing $N_\text{init}$ samples from $U(\mathcal{Z})$
\FOR{$Z_\text{all}\in\mathcal{D}_C$}
\STATE Compute $\text{PD}(\text{ACAP}^{-1}(Z_\text{all},\theta_D))$ and $I_c^*(Z_\text{all})$
\ENDFOR
\STATE Update $\theta_C$ using $\mathcal{D}_C$
\WHILE{Not converged}
\STATE{${\Delta\mathcal{D}_C}_1\gets$Draw $N_\text{aug}/2$ samples from $\mathcal{D}_C$} 
\FOR{$Z_\text{all}\in{\Delta\mathcal{D}_C}_1$}
\STATE{$Z_\text{all}^\text{last}\gets Z_\text{all}$}
\WHILE{True}
\STATE Update $Z_\text{all}$ using \prettyref{eq:newton}
\IF{$\|Z_\text{all}-Z_\text{all}^\text{last}\|_\infty$ sufficiently small}
\STATE Break
\ENDIF
\STATE{$Z_\text{all}^\text{last}\gets Z_\text{all}$}
\ENDWHILE
\STATE $\mathcal{D}_C\gets\mathcal{D}_C\bigcup\{\left<Z_\text{all}^\text{last},I_c^*(Z_\text{all}^\text{last})\right>\}$
\ENDFOR
\STATE{${\Delta\mathcal{D}_C}_2\gets$Draw $N_\text{aug}/2$ samples from $U(\mathcal{Z})$}
\FOR{$Z_\text{all}\in{\Delta\mathcal{D}_C}_2$}
\STATE $\mathcal{D}_C\gets\mathcal{D}_C\bigcup\{\left<Z_\text{all},I_c^*(Z_\text{all})\right>\}$
\ENDFOR
\STATE Update $\theta_C$ using $\mathcal{D}_C$
\ENDWHILE
\end{footnotesize}
\end{algorithmic}
\end{algorithm}

\begin{figure*}[t]
\vspace{-15px}
\centering
\scalebox{0.75}{
\arrayrulewidth=2pt
\def\arraystretch{1.5}
\setlength{\tabcolsep}{0pt}
\begin{tabular}{c?c?c?c?c?c?c?}
\hhline{~|-----}
&
\FIGTITLE&
\FIGTITLE&
\FIGTITLE&
\FIGTITLE&
\FIGTITLE\\
\hline
\multicolumn{1}{?c?}{(a)}&
\includegraphics[width=0.19\linewidth]{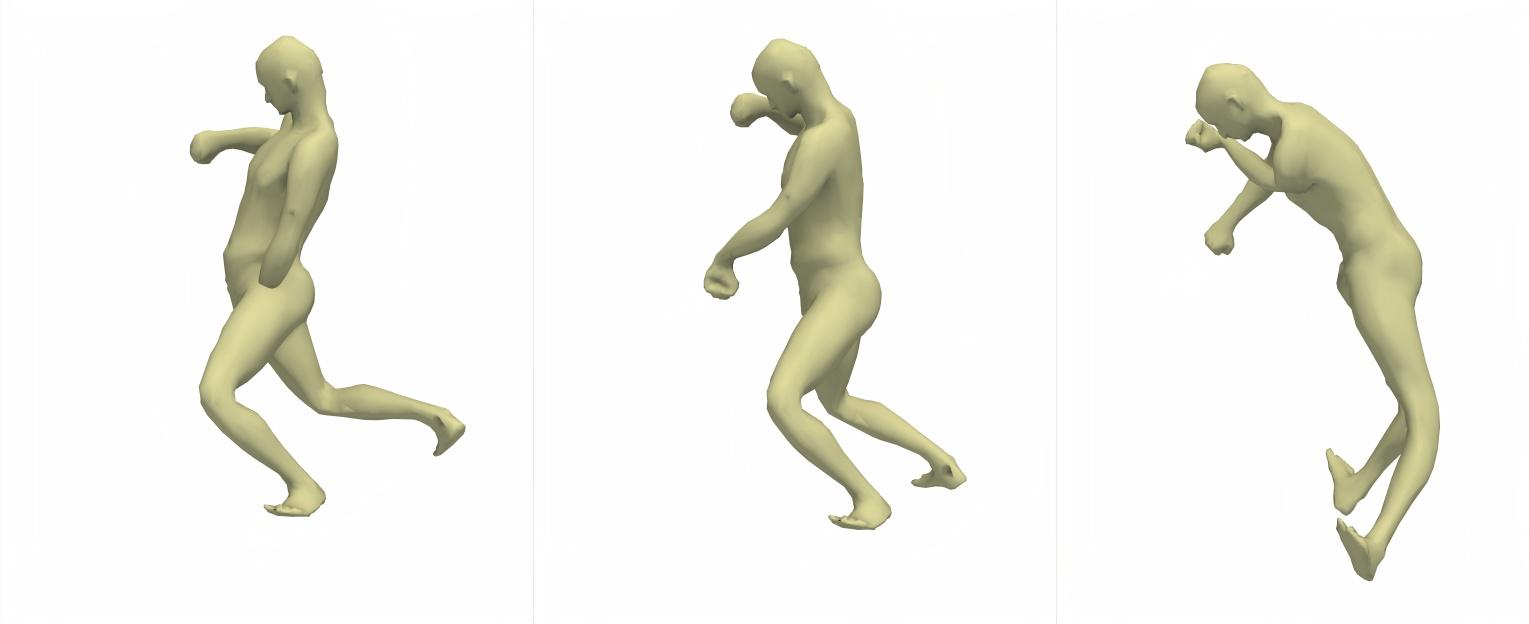} &
\includegraphics[width=0.19\linewidth]{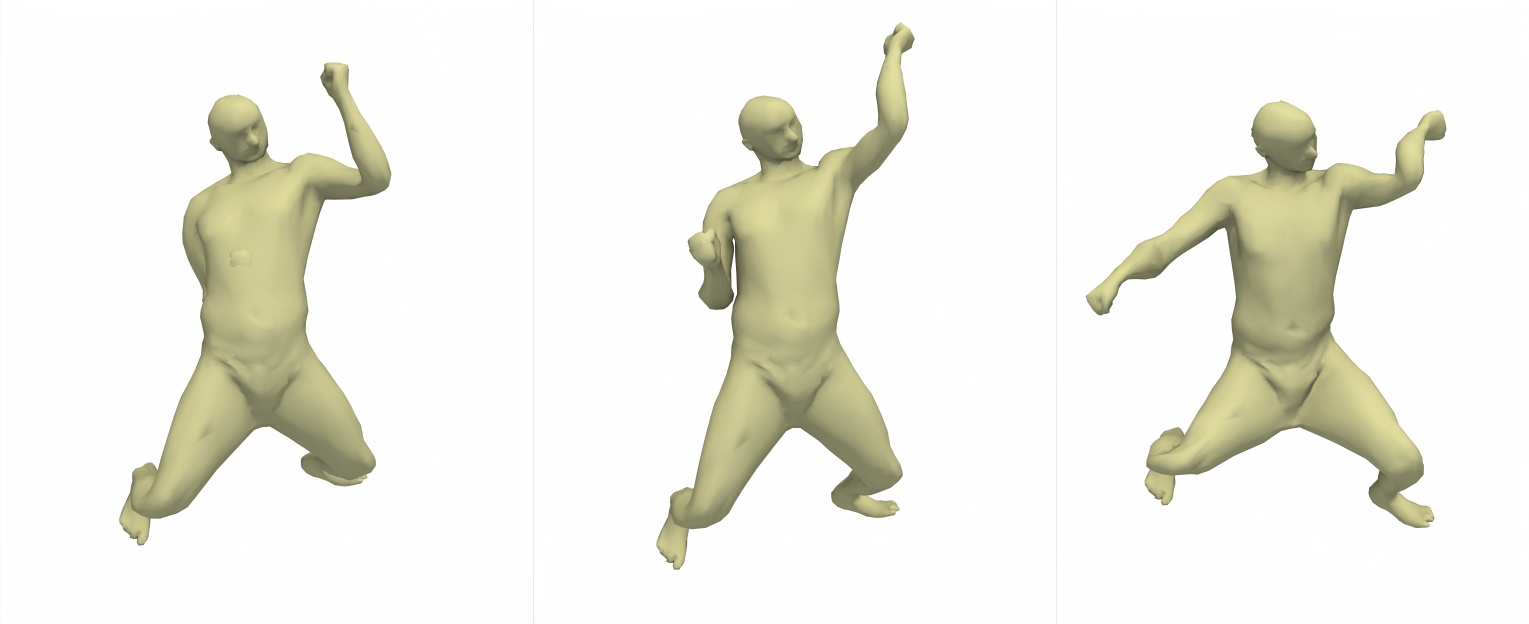} &
\includegraphics[width=0.19\linewidth]{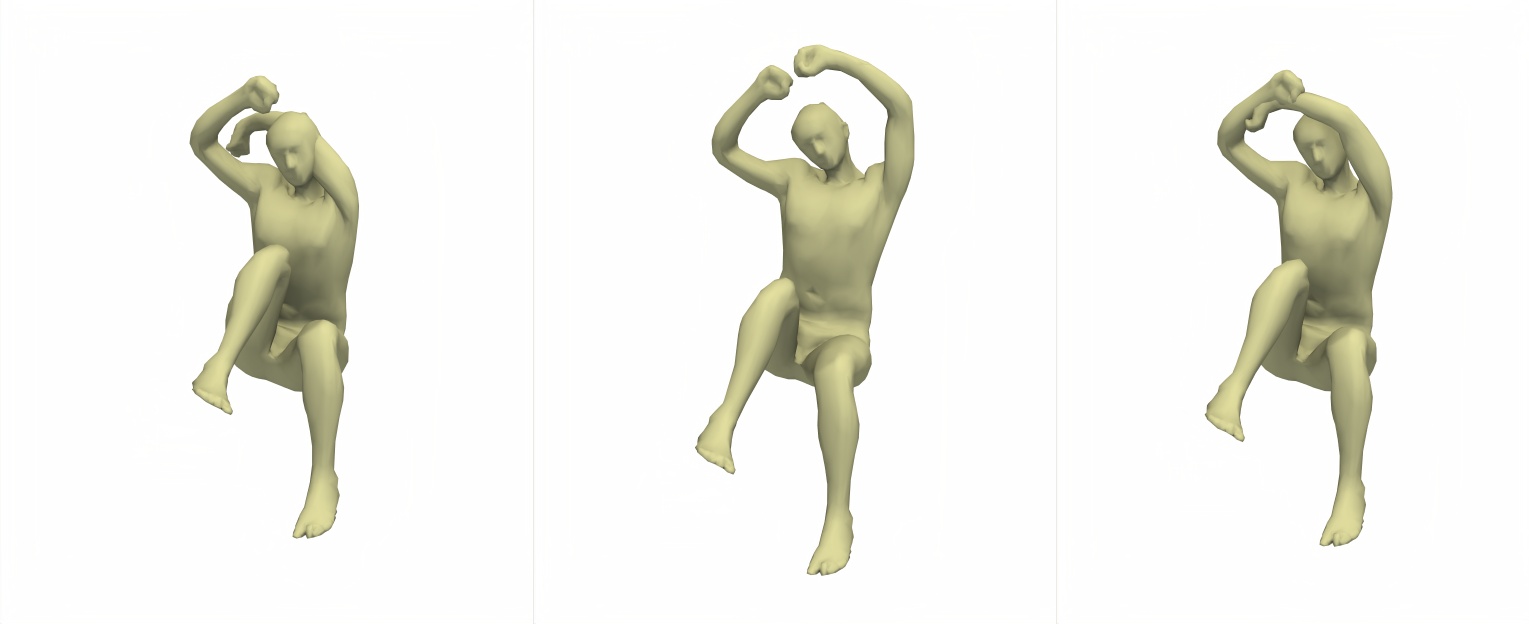} &
\includegraphics[width=0.19\linewidth]{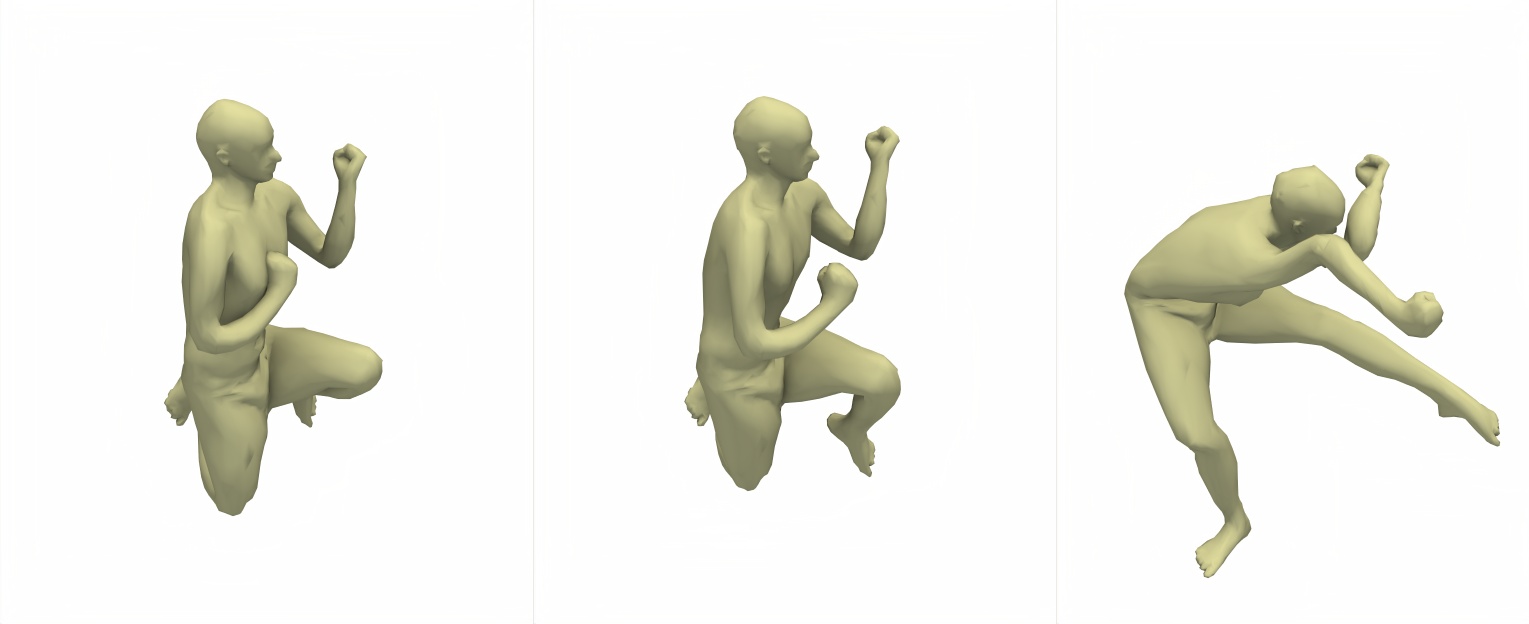} &
\includegraphics[width=0.19\linewidth]{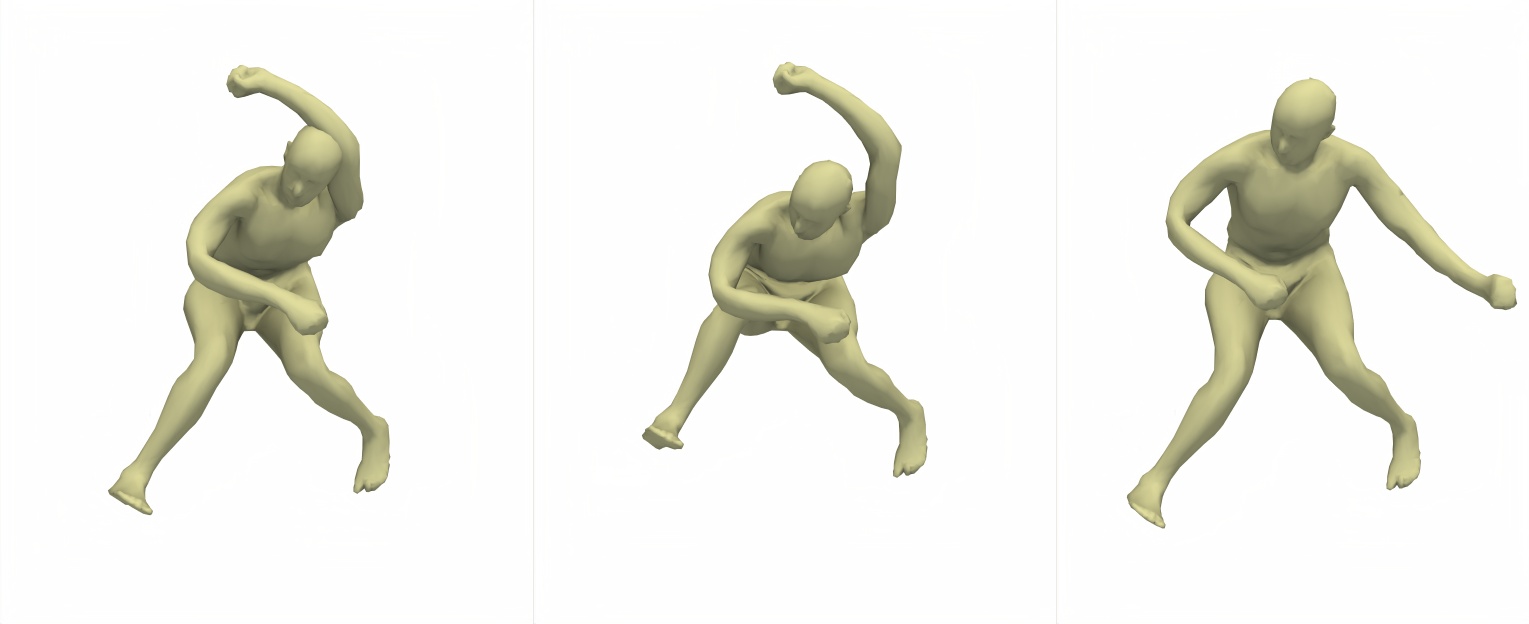}\\
\multicolumn{1}{?c?}{(b)}&
\includegraphics[width=0.19\linewidth]{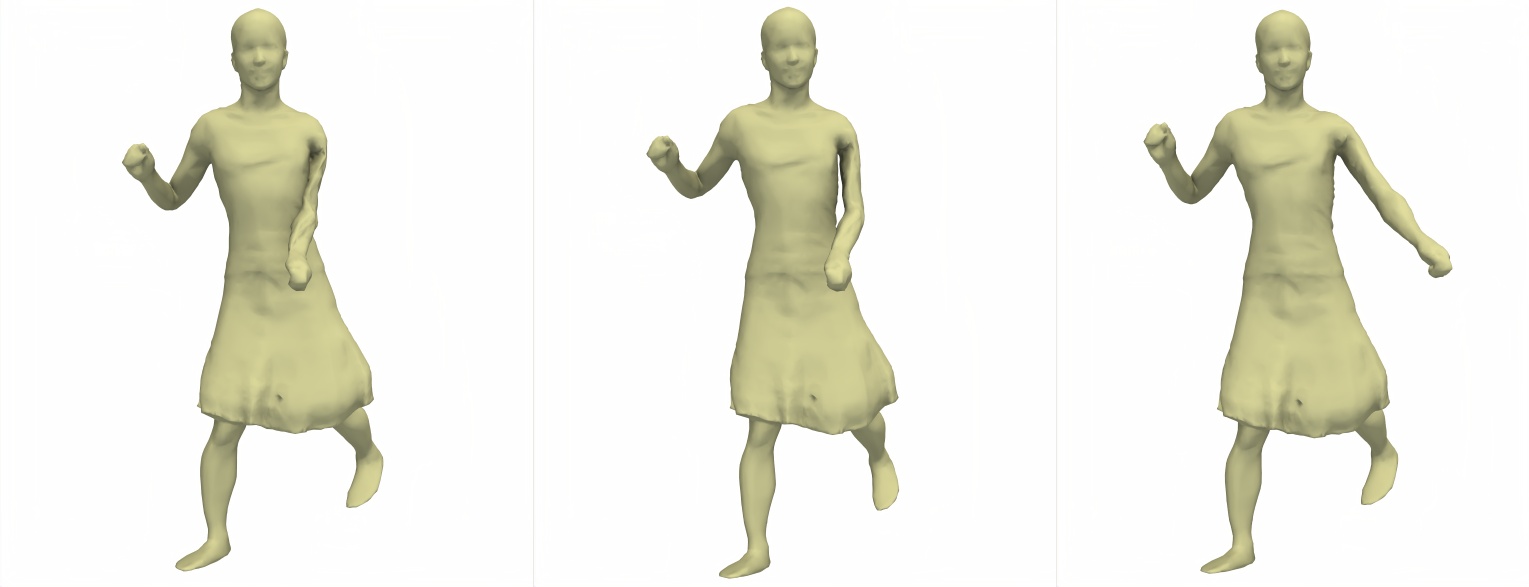} &
\includegraphics[width=0.19\linewidth]{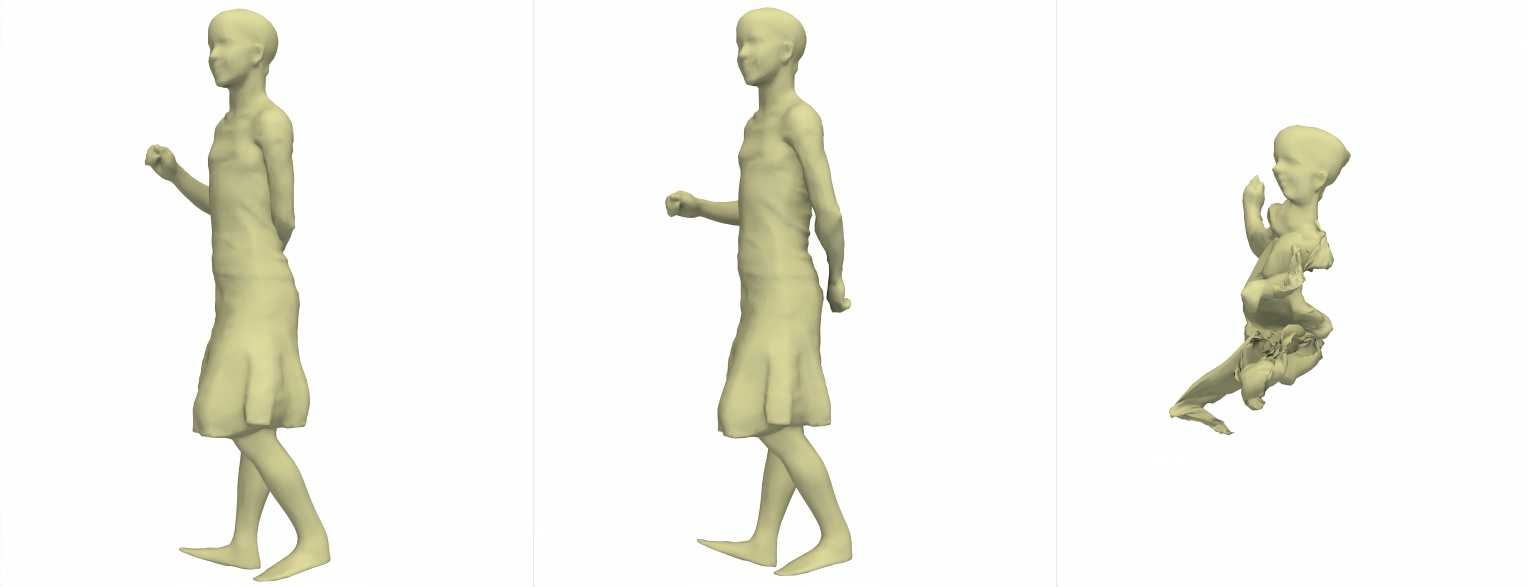} &
\includegraphics[width=0.19\linewidth]{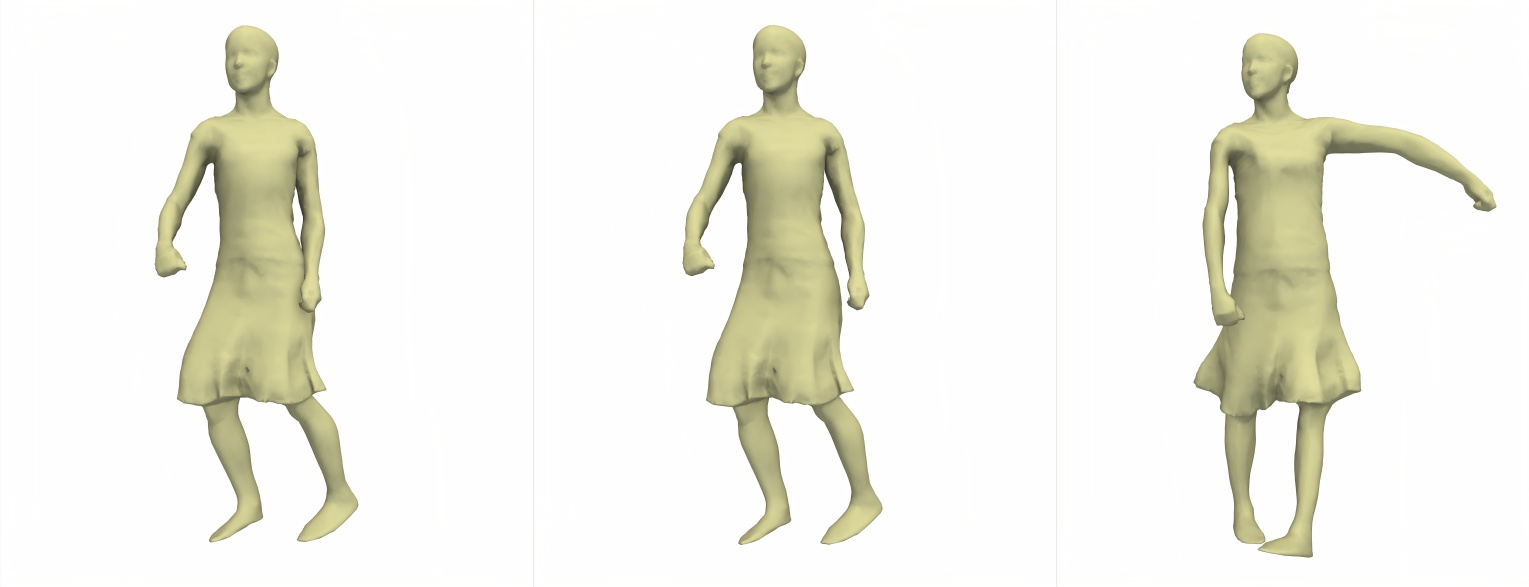} &
\includegraphics[width=0.19\linewidth]{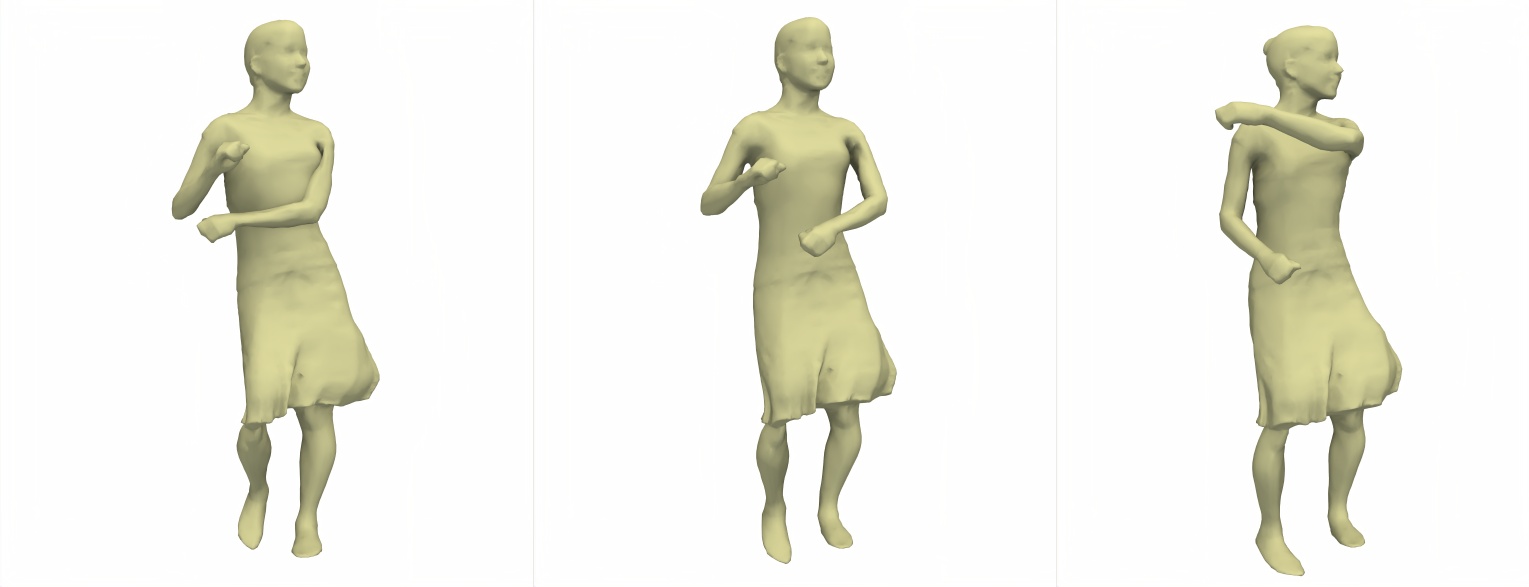} &
\includegraphics[width=0.19\linewidth]{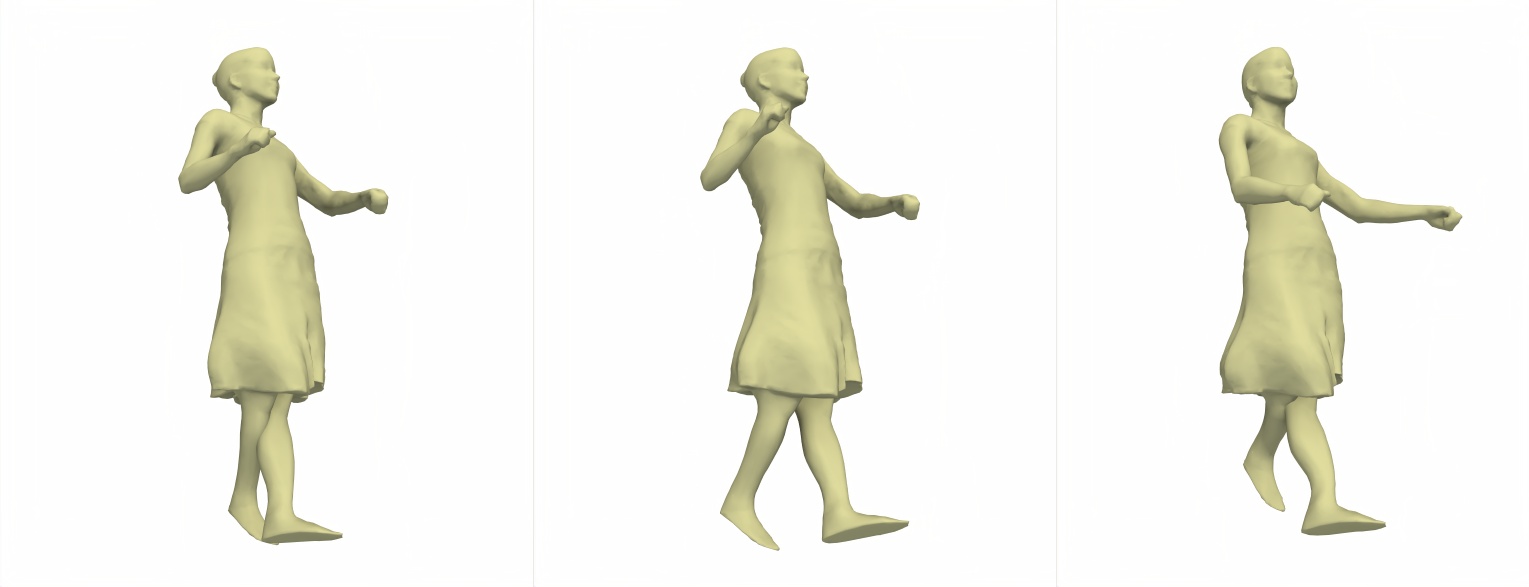}\\
\multicolumn{1}{?c?}{(c)}&
\includegraphics[width=0.19\linewidth]{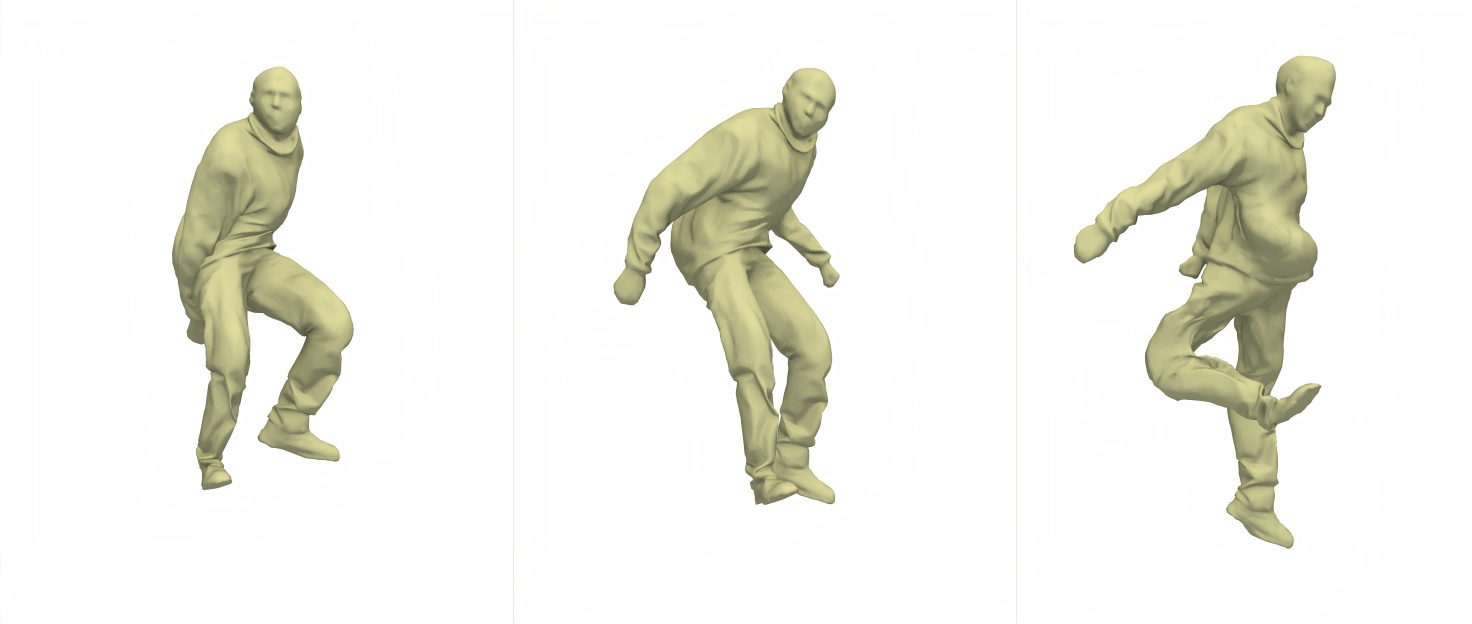} &
\includegraphics[width=0.19\linewidth]{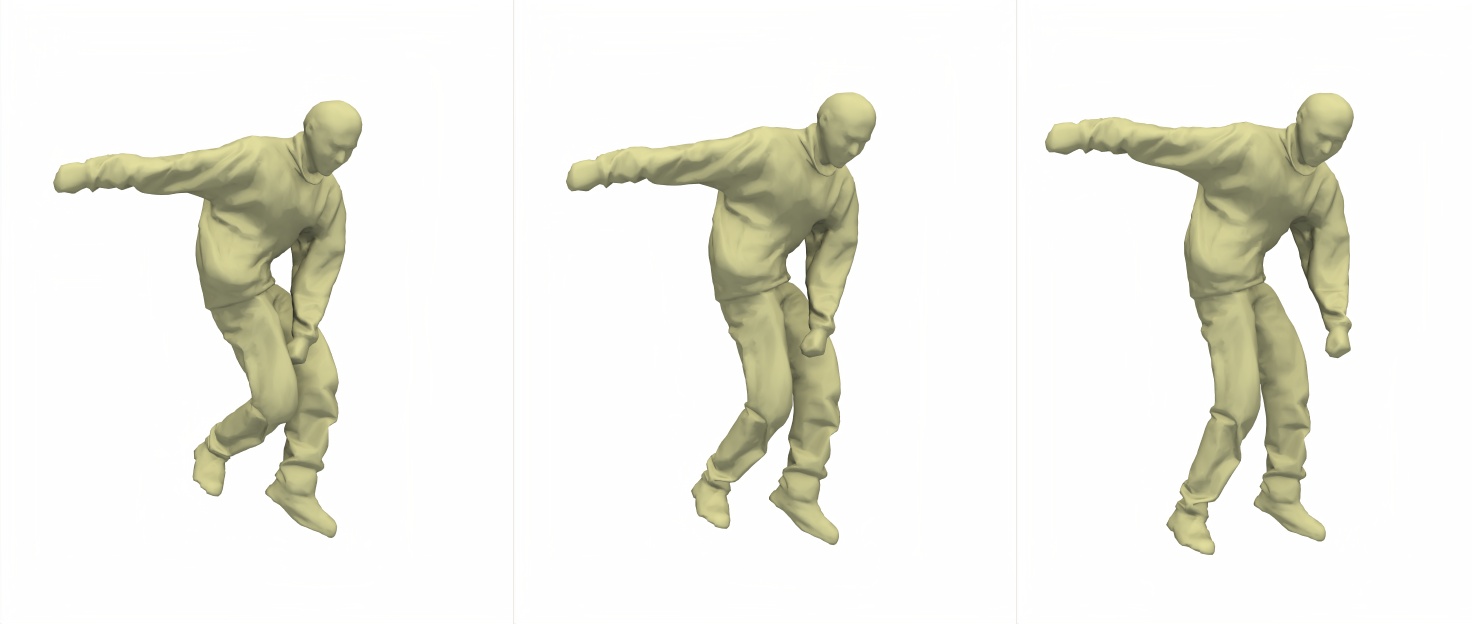} &
\includegraphics[width=0.19\linewidth]{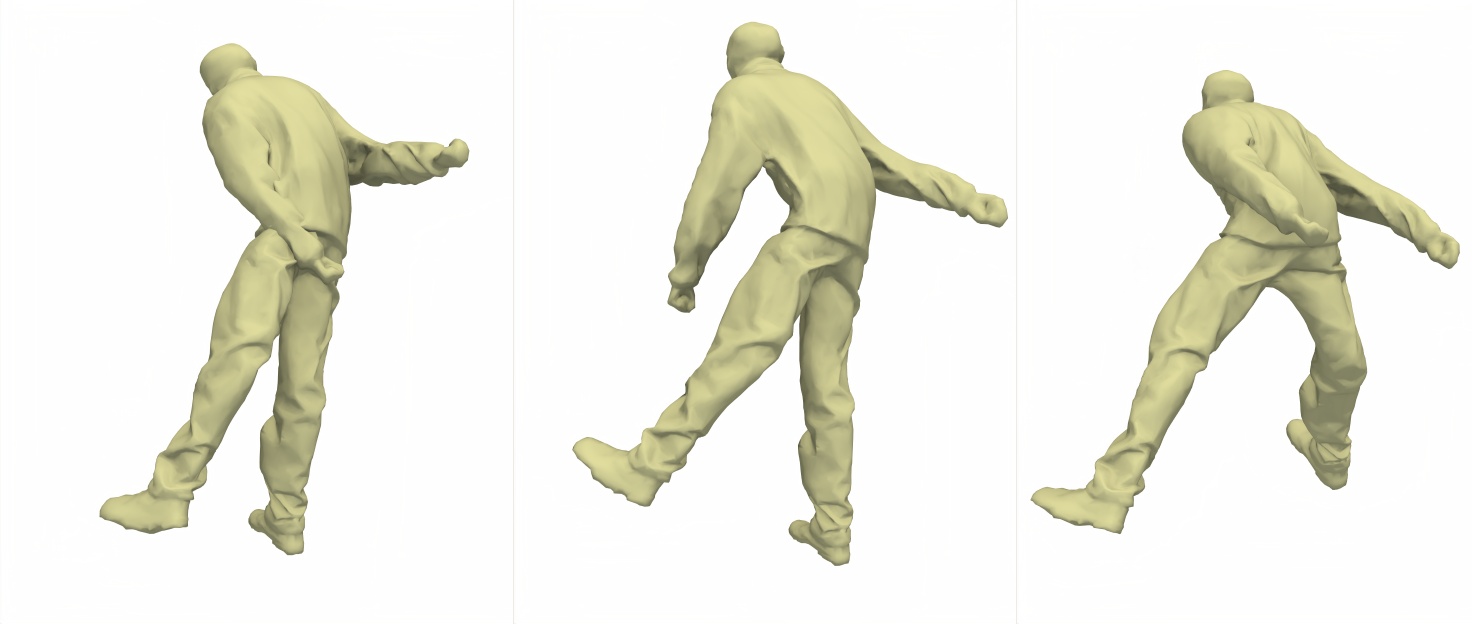} &
\includegraphics[width=0.19\linewidth]{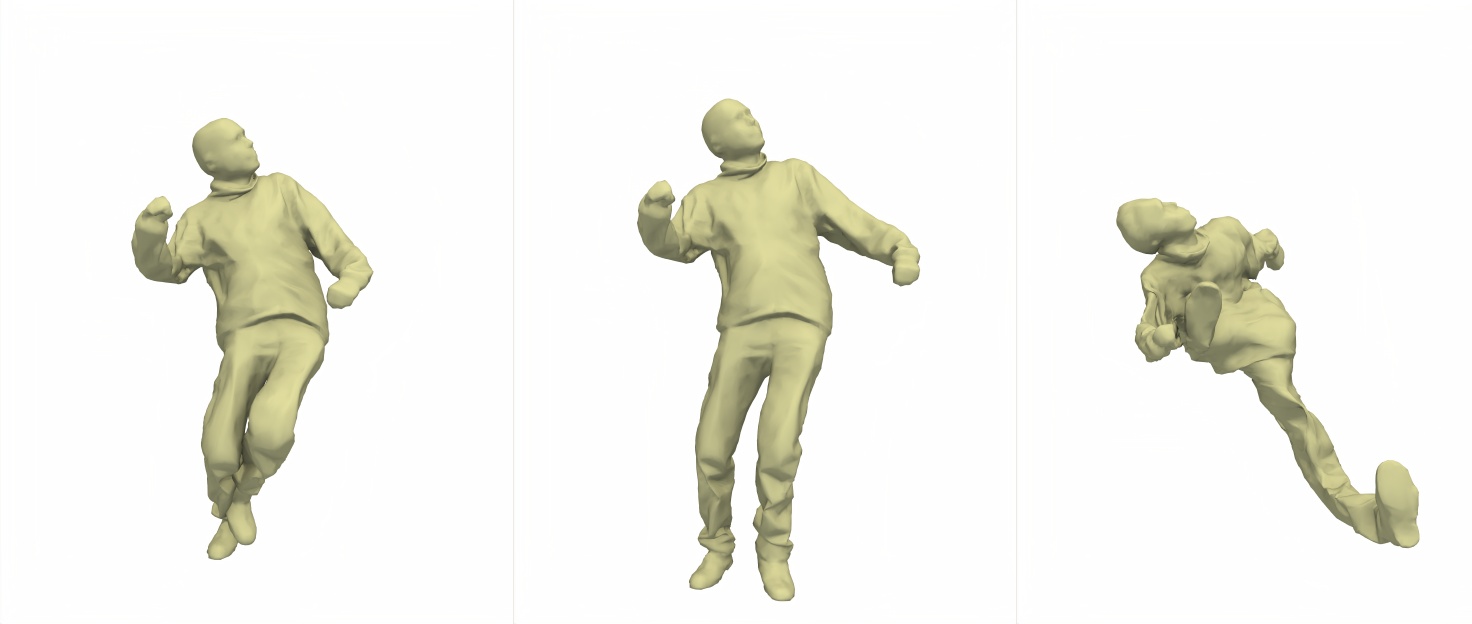} &
\includegraphics[width=0.19\linewidth]{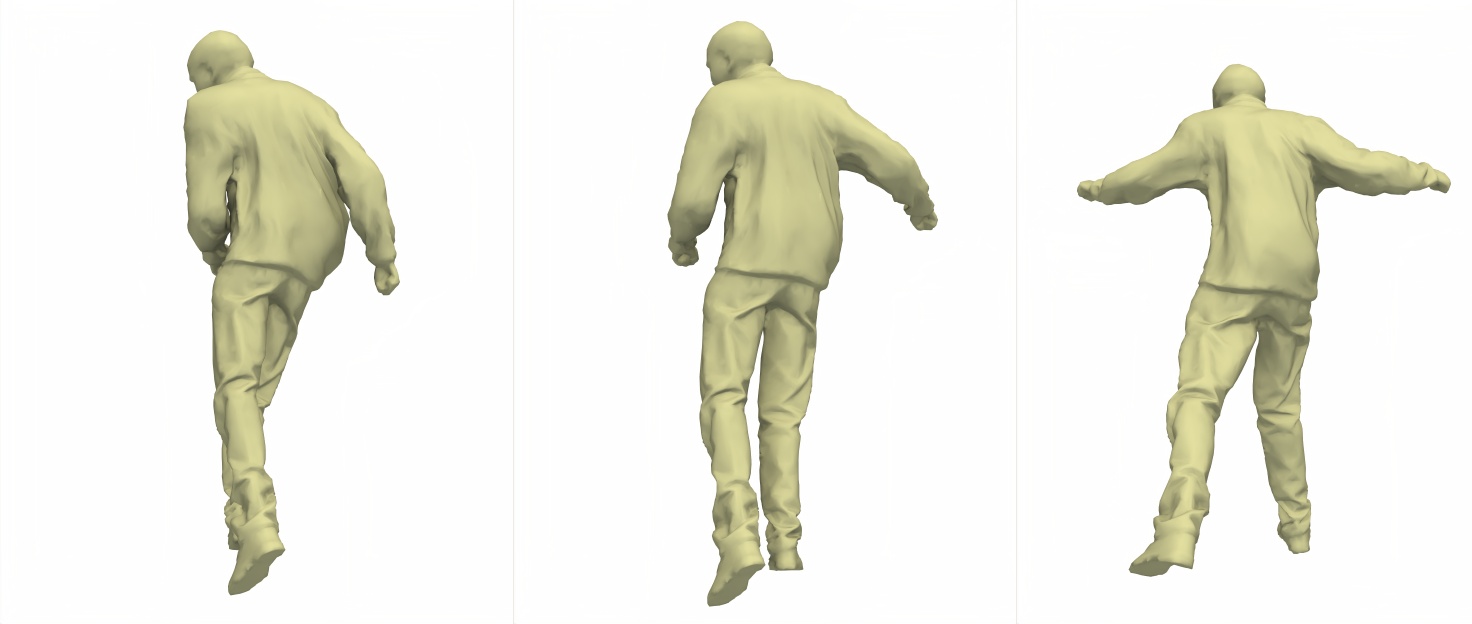}\\
\multicolumn{1}{?c?}{(d)}&
\includegraphics[width=0.19\linewidth]{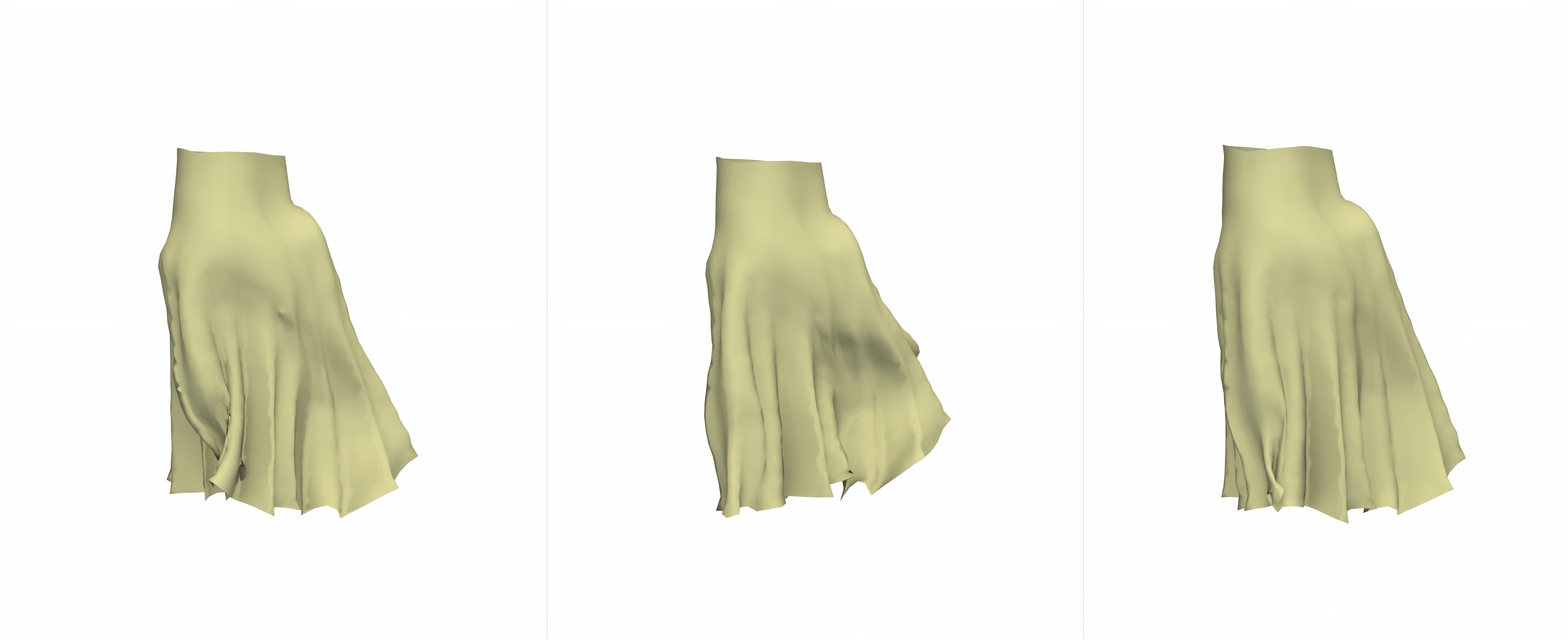} &
\includegraphics[width=0.19\linewidth]{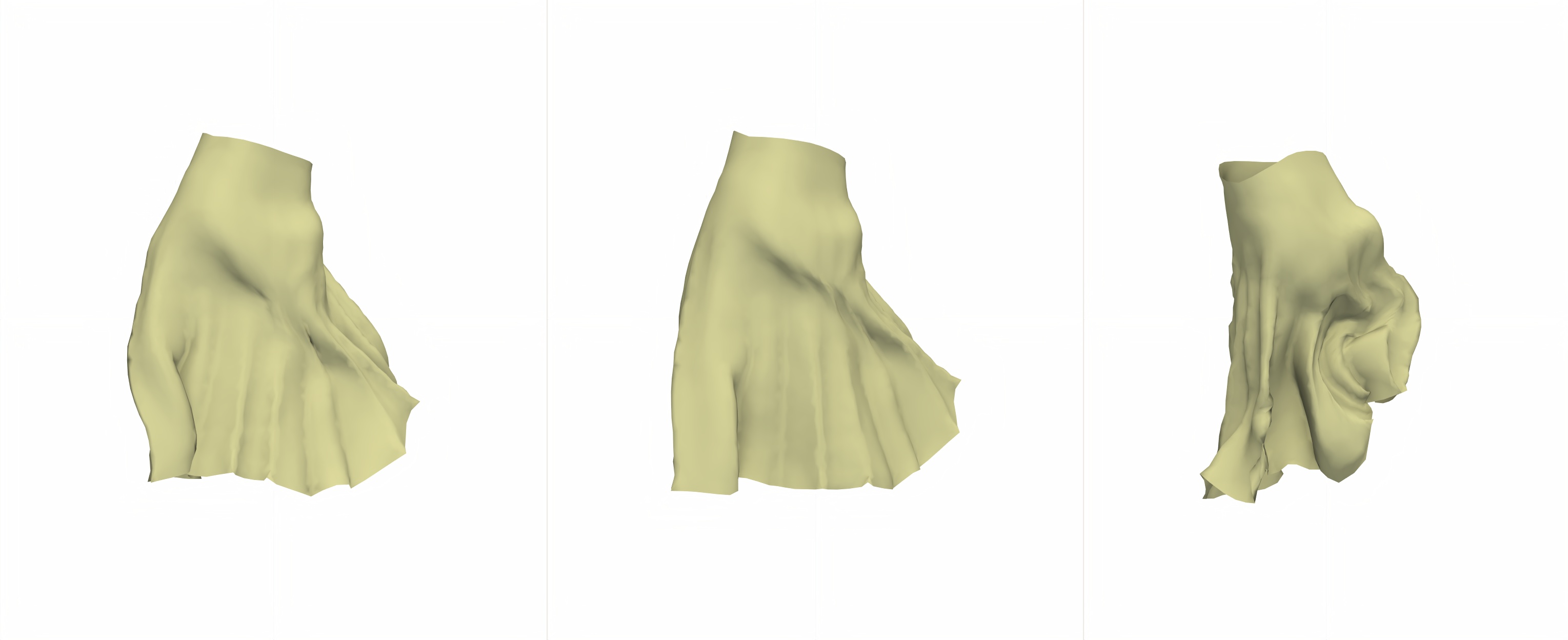} &
\includegraphics[width=0.19\linewidth]{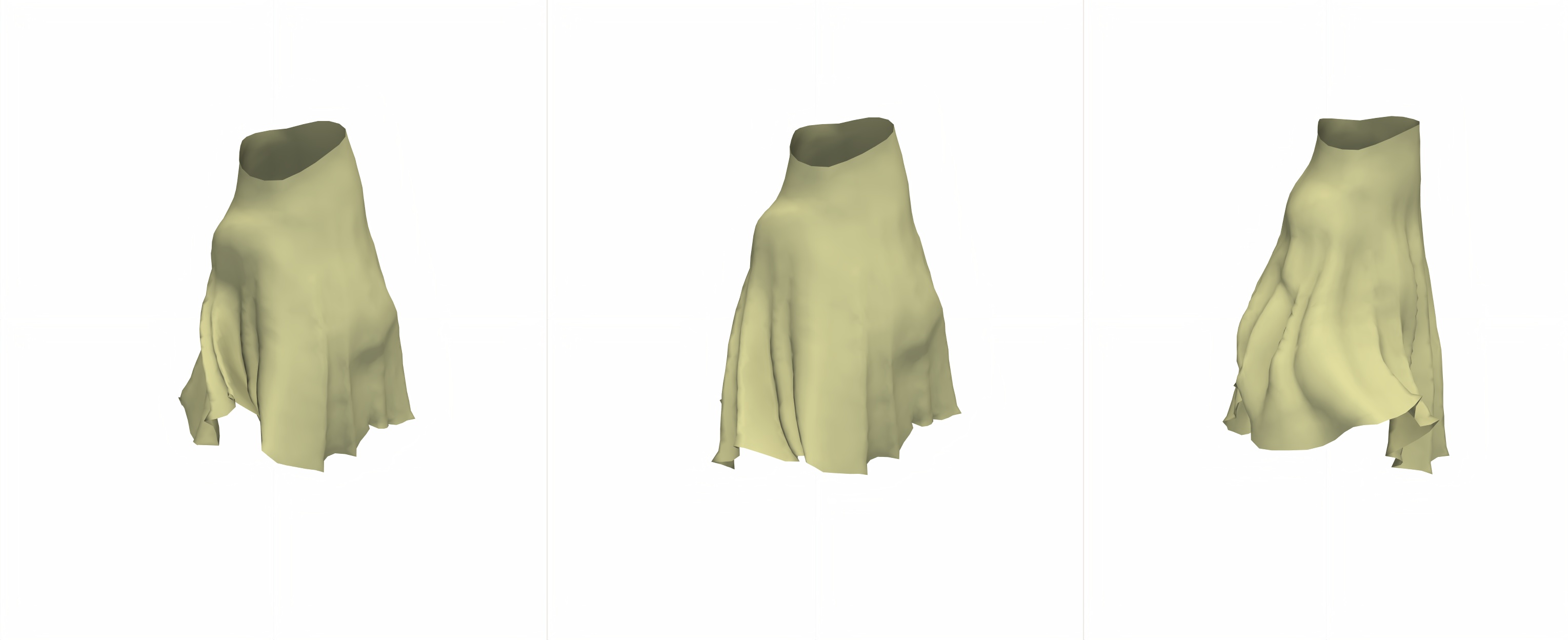} &
\includegraphics[width=0.19\linewidth]{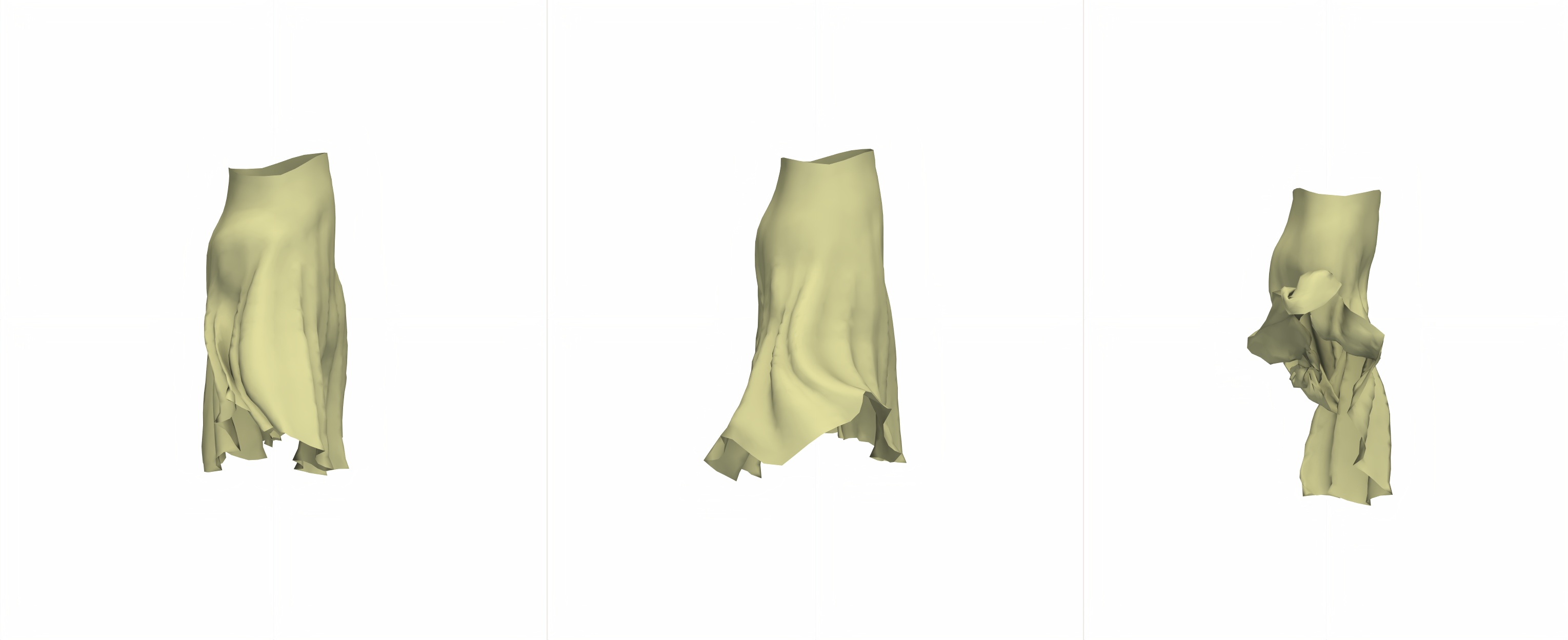} &
\includegraphics[width=0.19\linewidth]{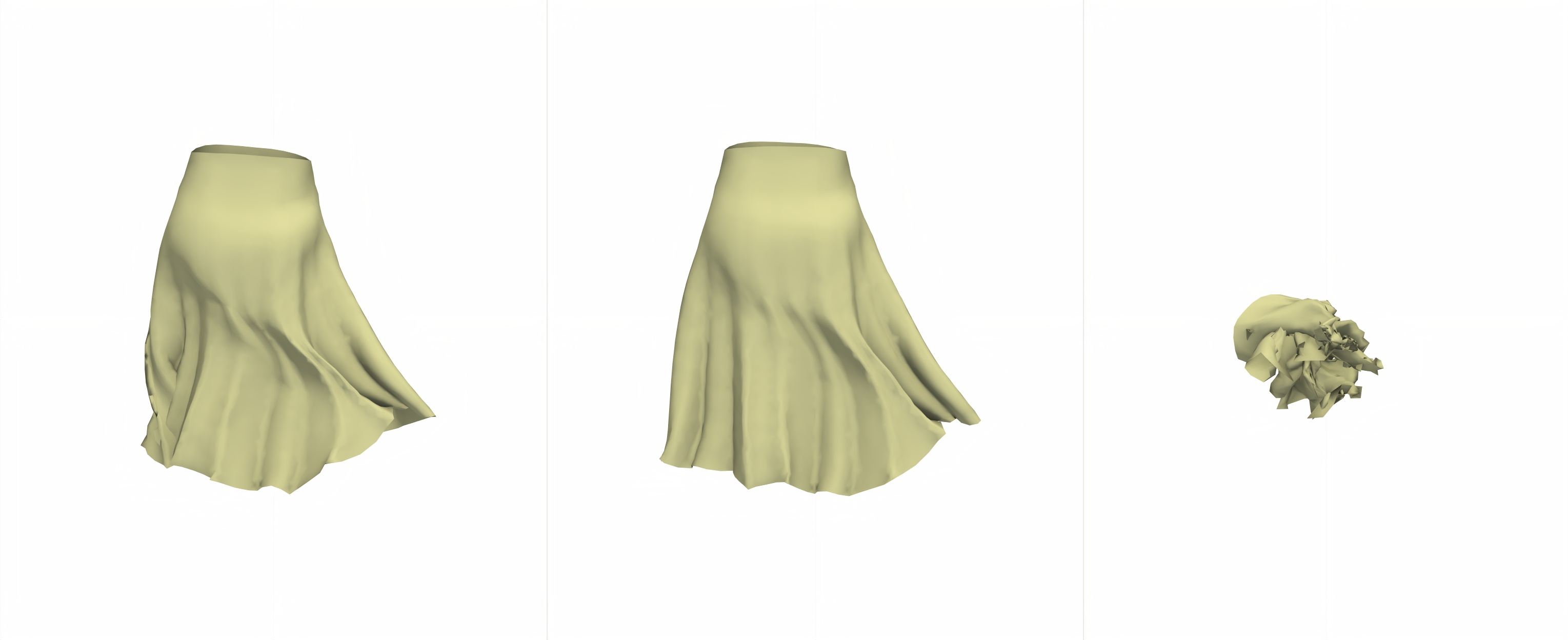}\\
\multicolumn{1}{?c?}{(e)}&
\includegraphics[width=0.19\linewidth]{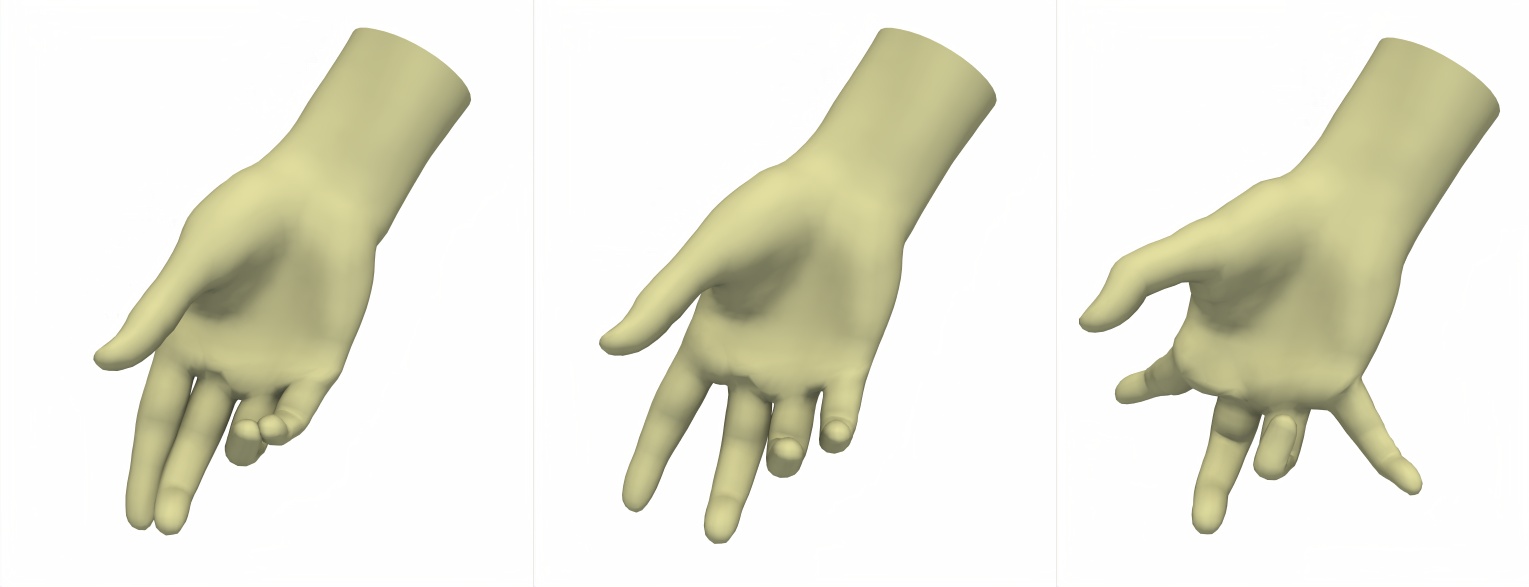} &
\includegraphics[width=0.19\linewidth]{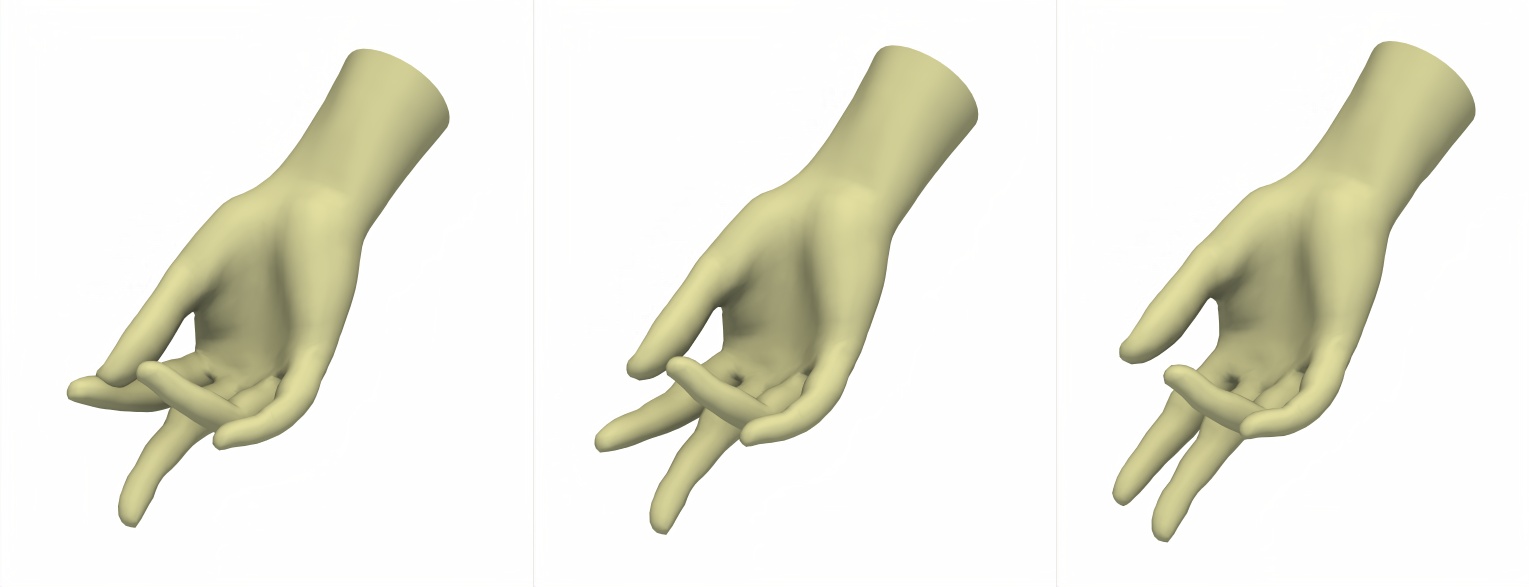} &
\includegraphics[width=0.19\linewidth]{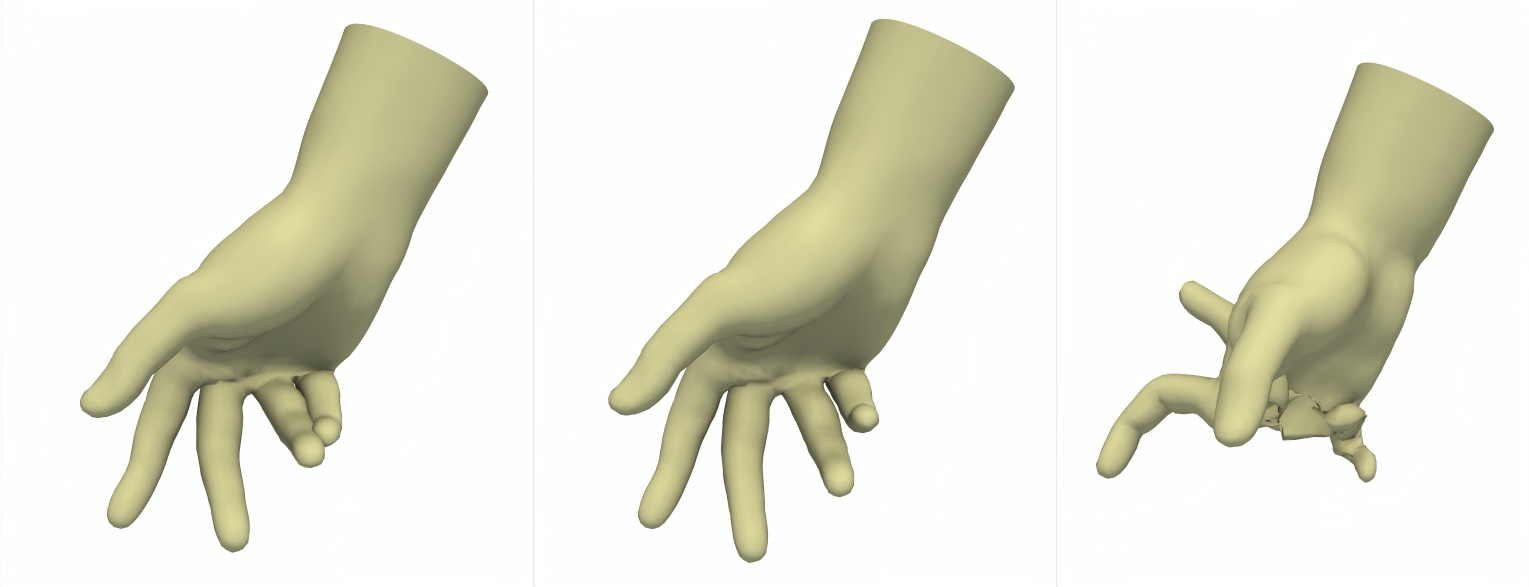} &
\includegraphics[width=0.19\linewidth]{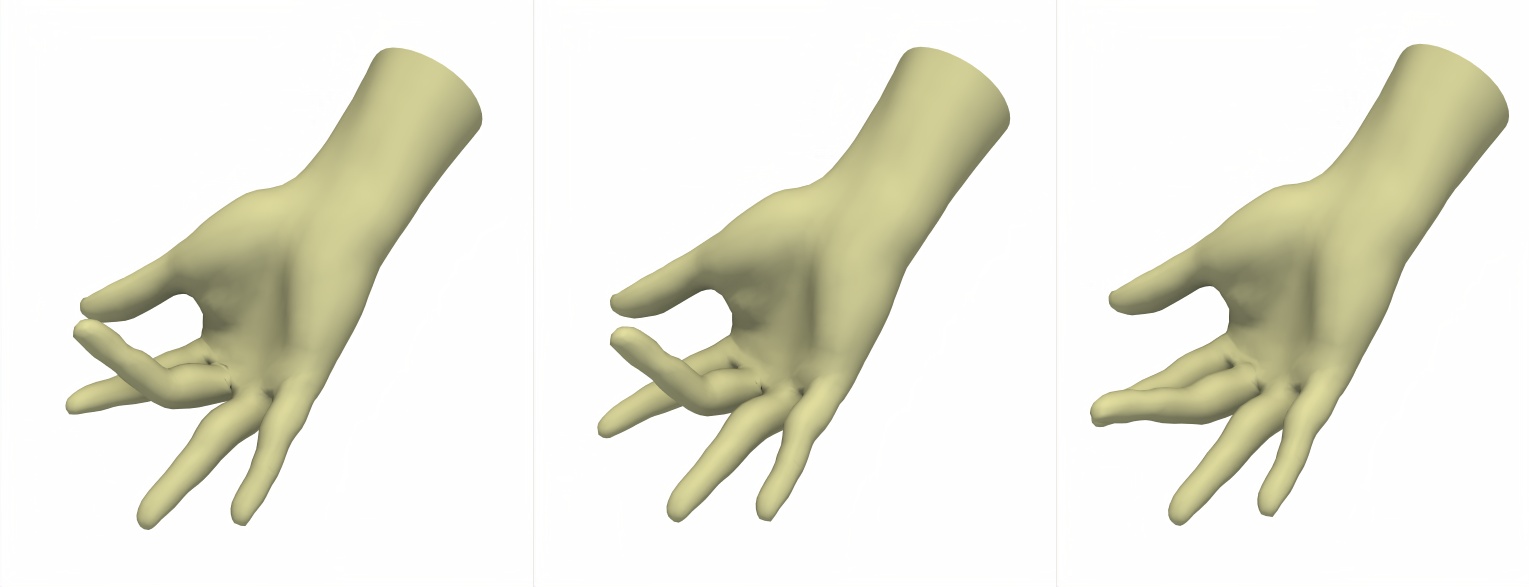} &
\includegraphics[width=0.19\linewidth]{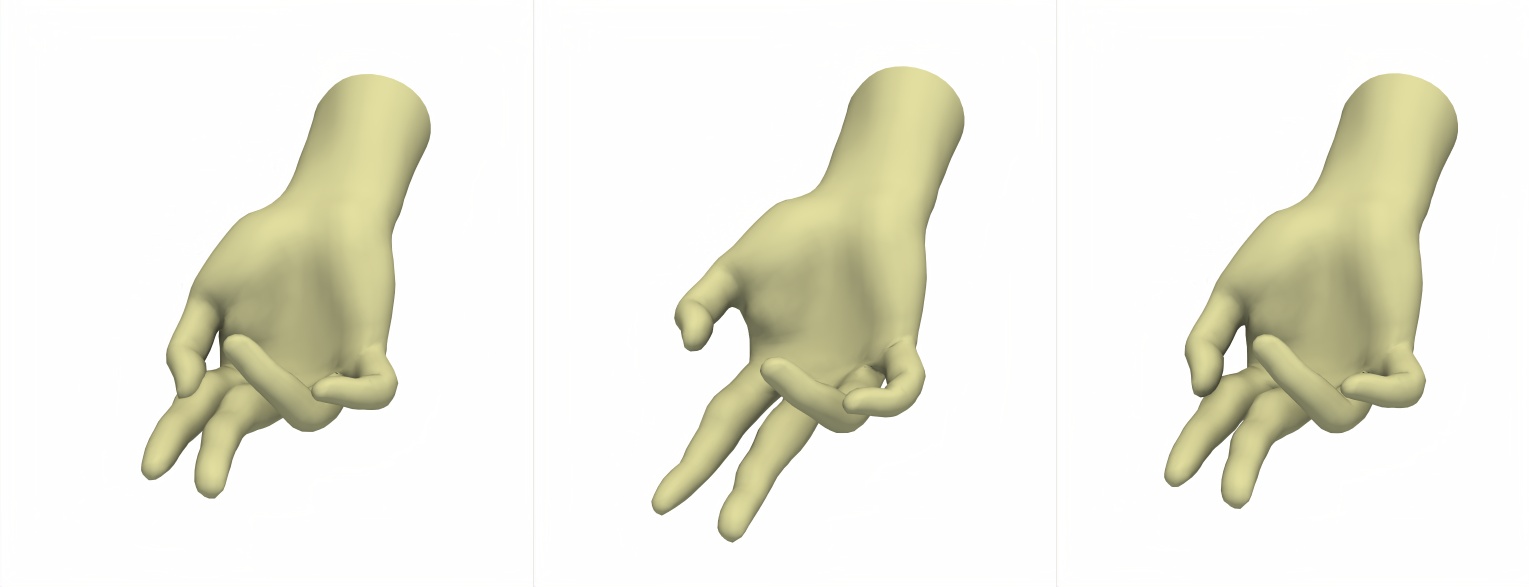}\\
\hline
\end{tabular}}
\vspace{-5px}
\caption{\footnotesize{\label{fig:examples} Representative examples of collision handling on the five datasets: (a) SCAPE; (b) MIT Swing; (c) MIT Jump; (d) Skirt; (e) Hand. For each example, we show the given self-penetrating mesh (left), our result (middle, \textbf{\textit{Active+bd}}), and \textbf{\textit{Supv+bd}} (right).}}
\vspace{-10px}
\end{figure*}

\begin{figure*}[h]
\centering
\setlength{\tabcolsep}{2pt}
\begin{tabular}{ccccc}
\includegraphics[width=0.18\linewidth]{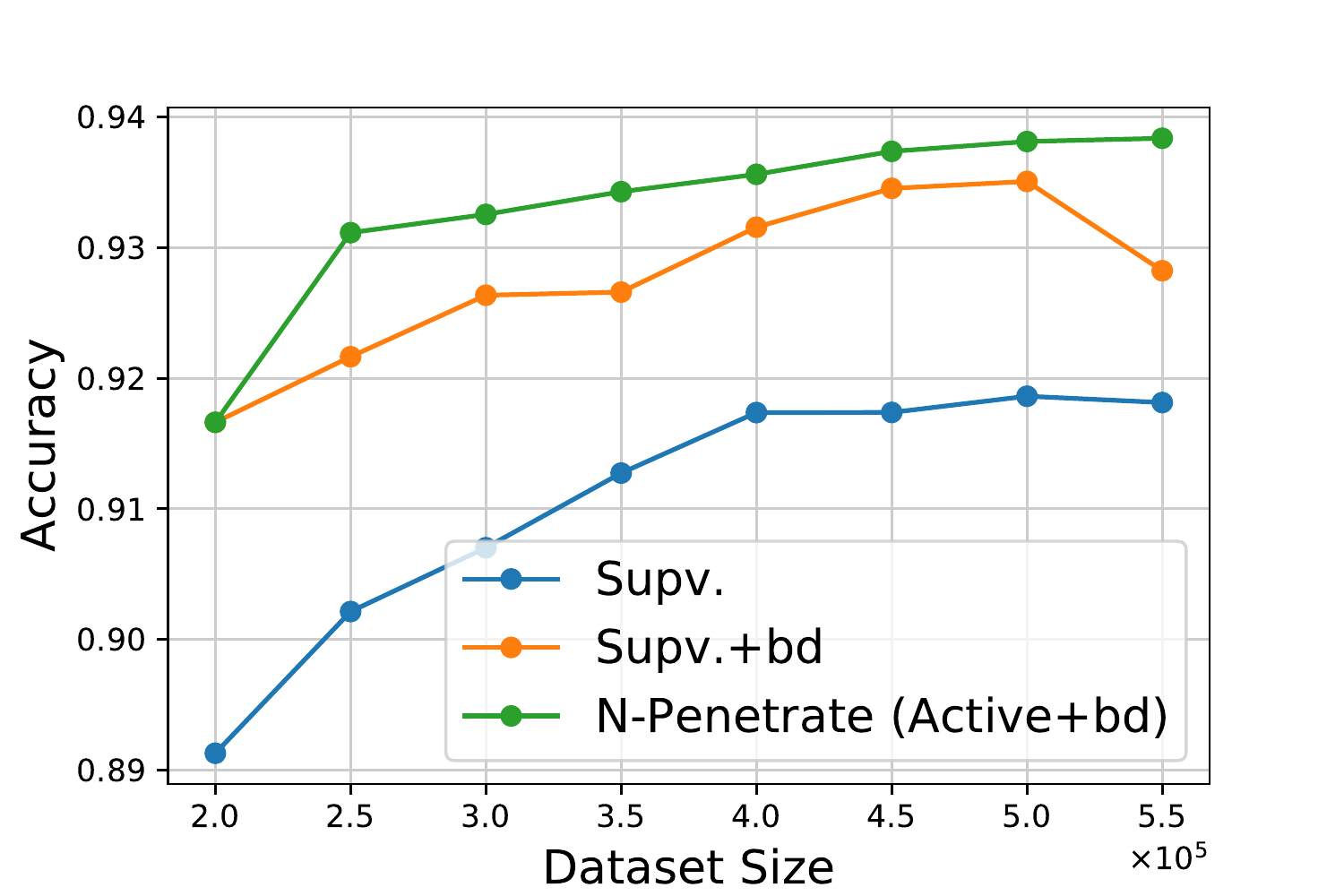}&
\includegraphics[width=0.18\linewidth]{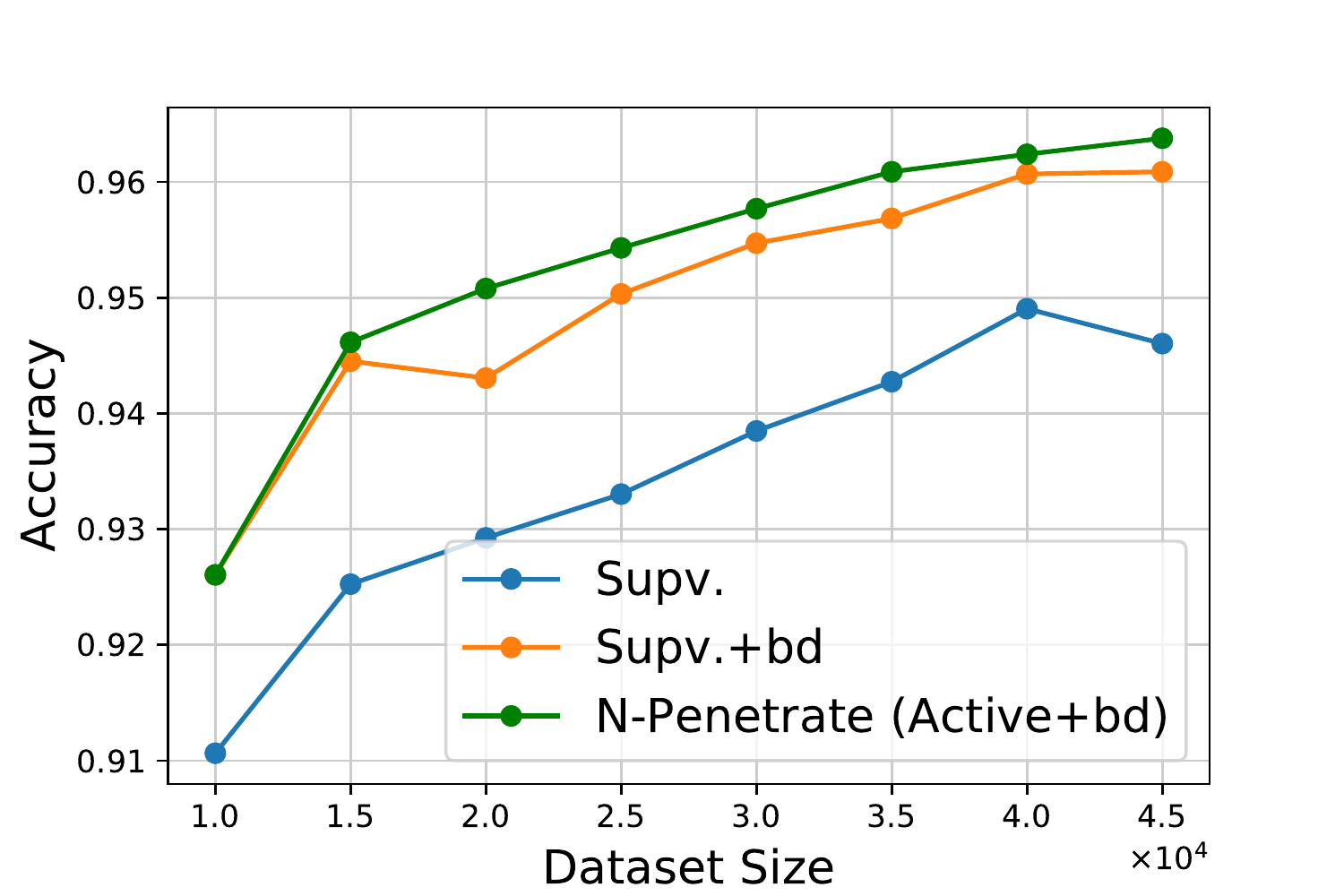}&
\includegraphics[width=0.18\linewidth]{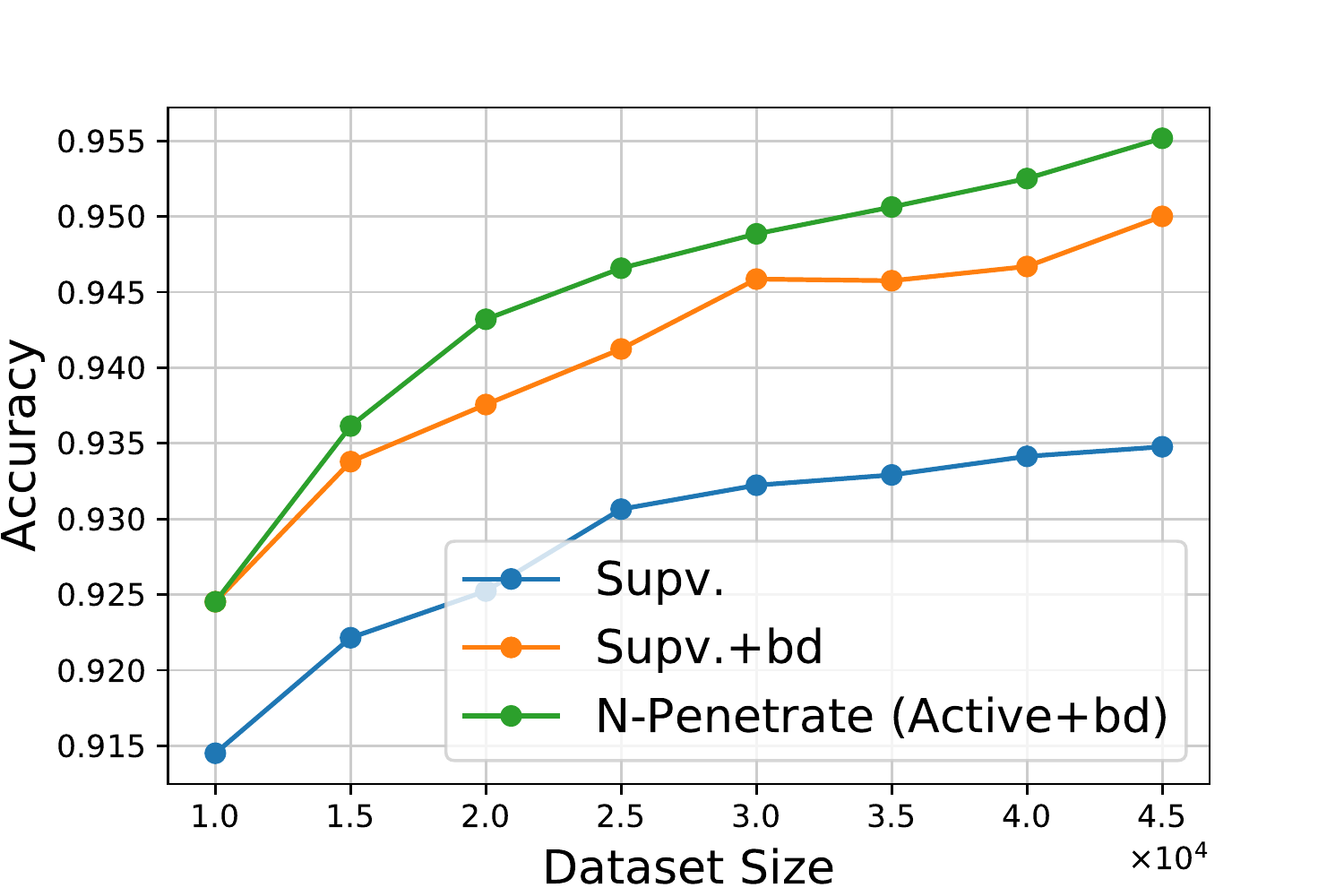}&
\includegraphics[width=0.18\linewidth]{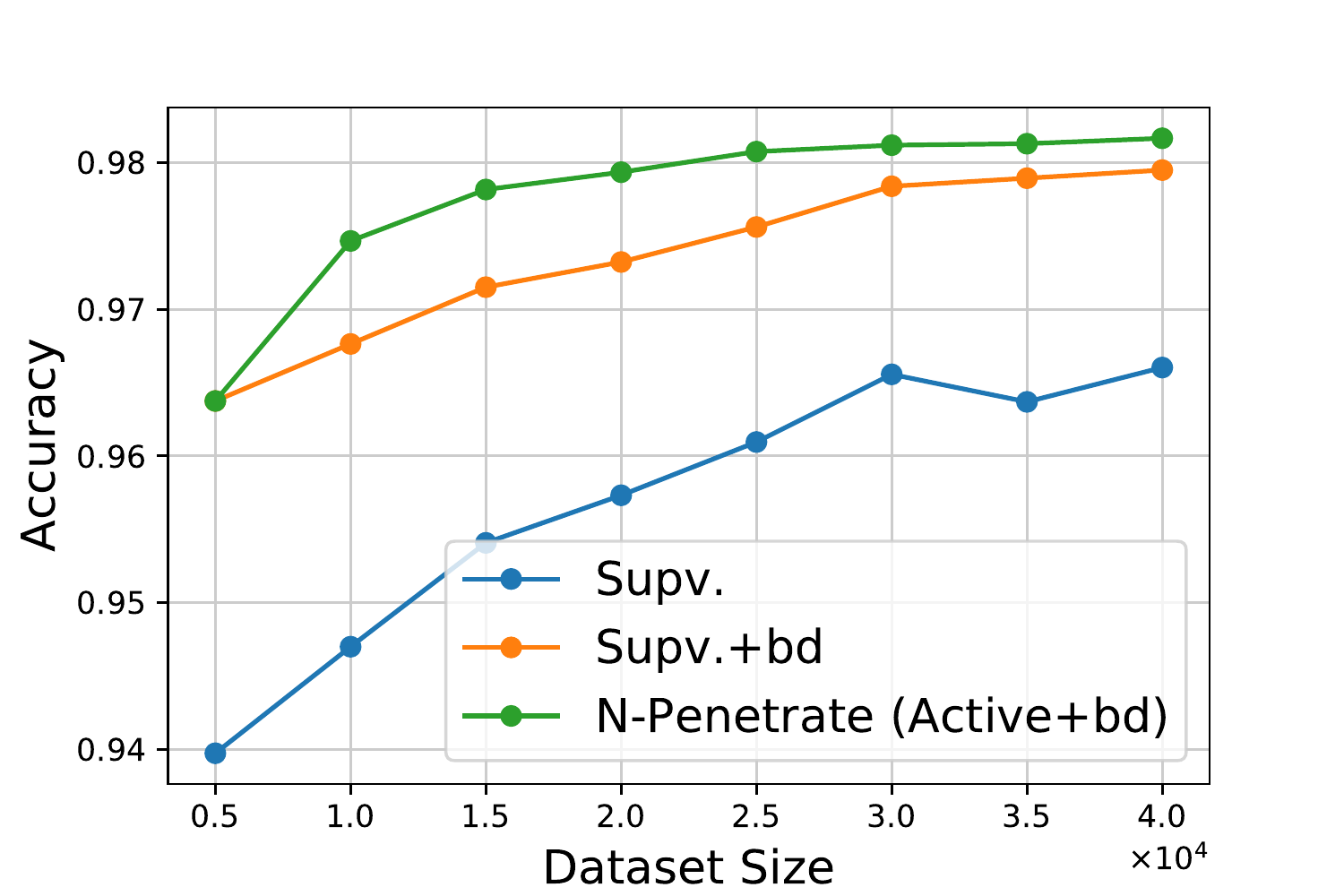}&
\includegraphics[width=0.18\linewidth]{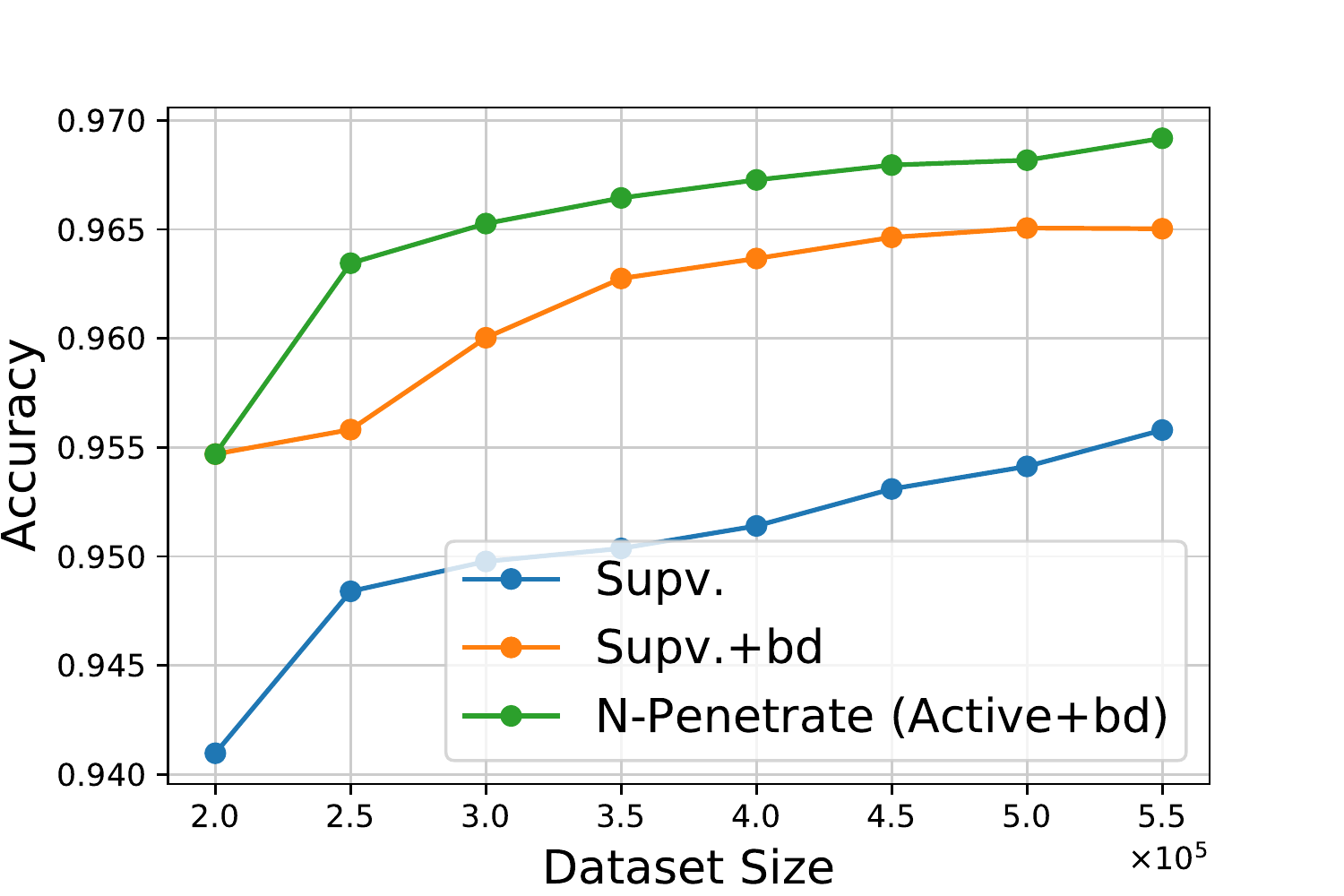}
\end{tabular}
\vspace{-10px}
\caption{\footnotesize{\label{fig:iterationAccuracy} We plot the accuracy of the neural collision detector against the dataset size. The baselines are trained using the same amount of data. On average, ours achieves $1.86\%$ higher accuracy than \textbf{\textit{Supv}}. From left to right: SCAPE, Swing, Jump, Skirt, and Hand.}}
\vspace{-10px}
\end{figure*}

\begin{figure*}[h]
\vspace{-15px}
\centering
\setlength{\tabcolsep}{2pt}
\begin{tabular}{ccccc}
\includegraphics[width=0.18\linewidth]{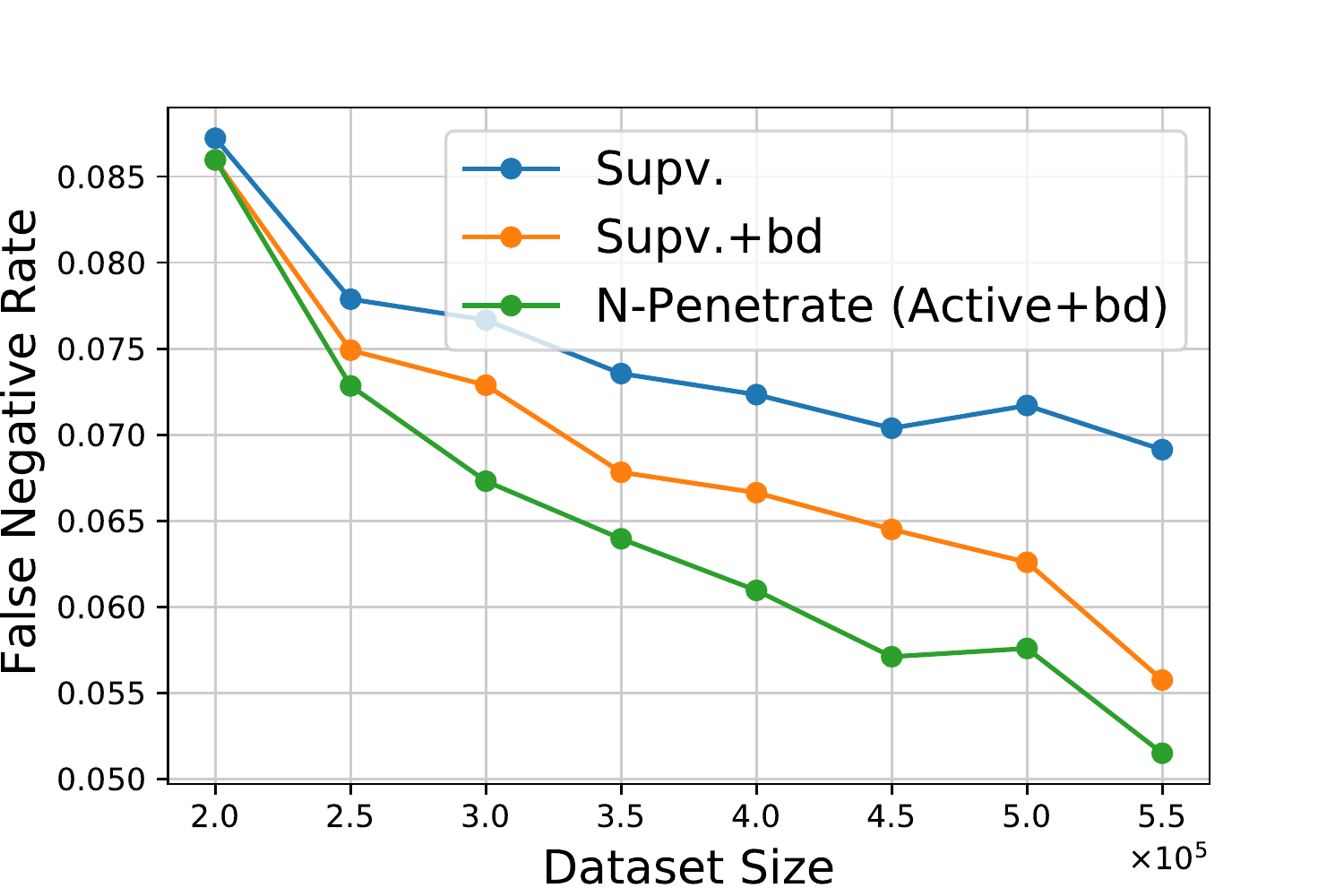}&
\includegraphics[width=0.18\linewidth]{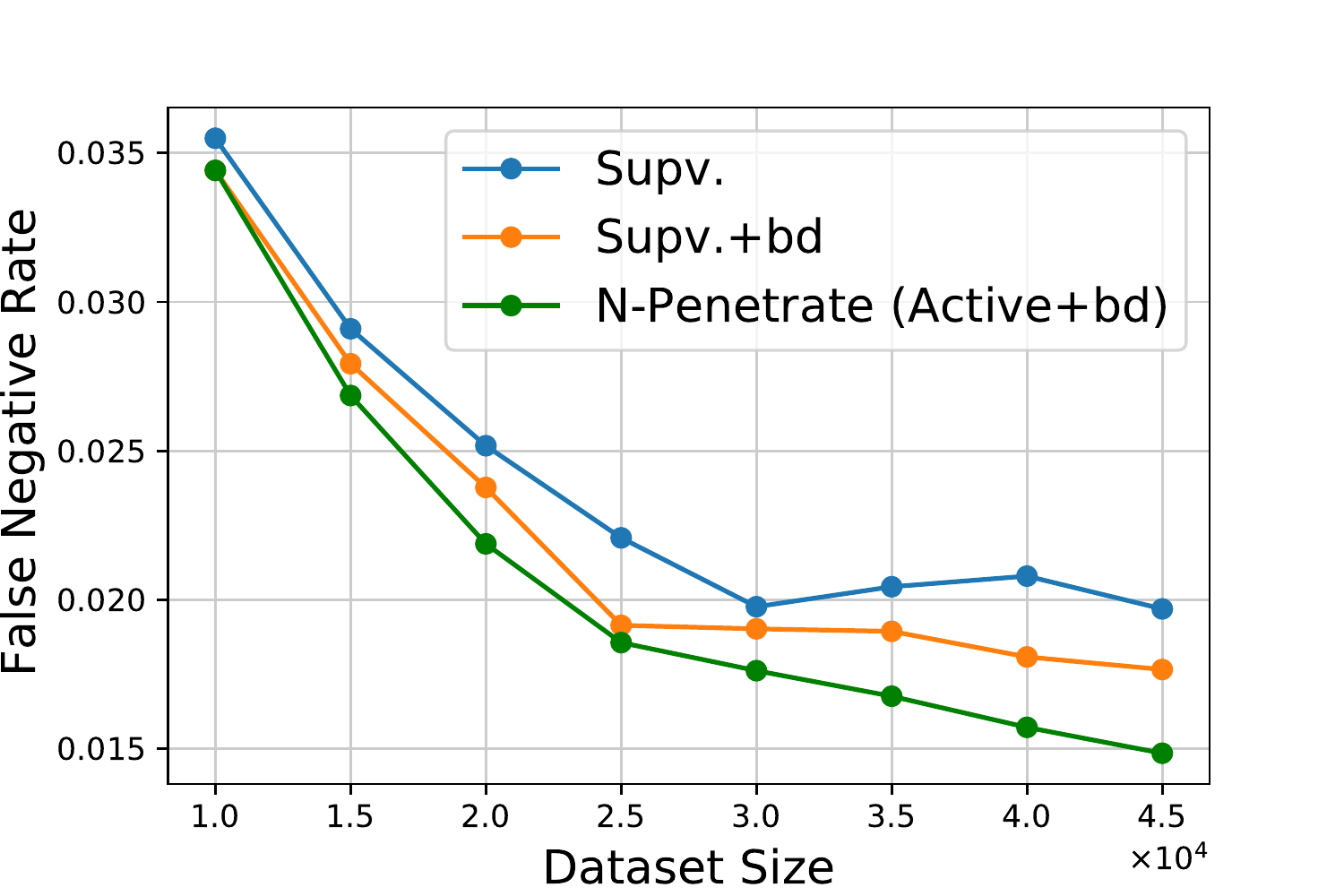}&
\includegraphics[width=0.18\linewidth]{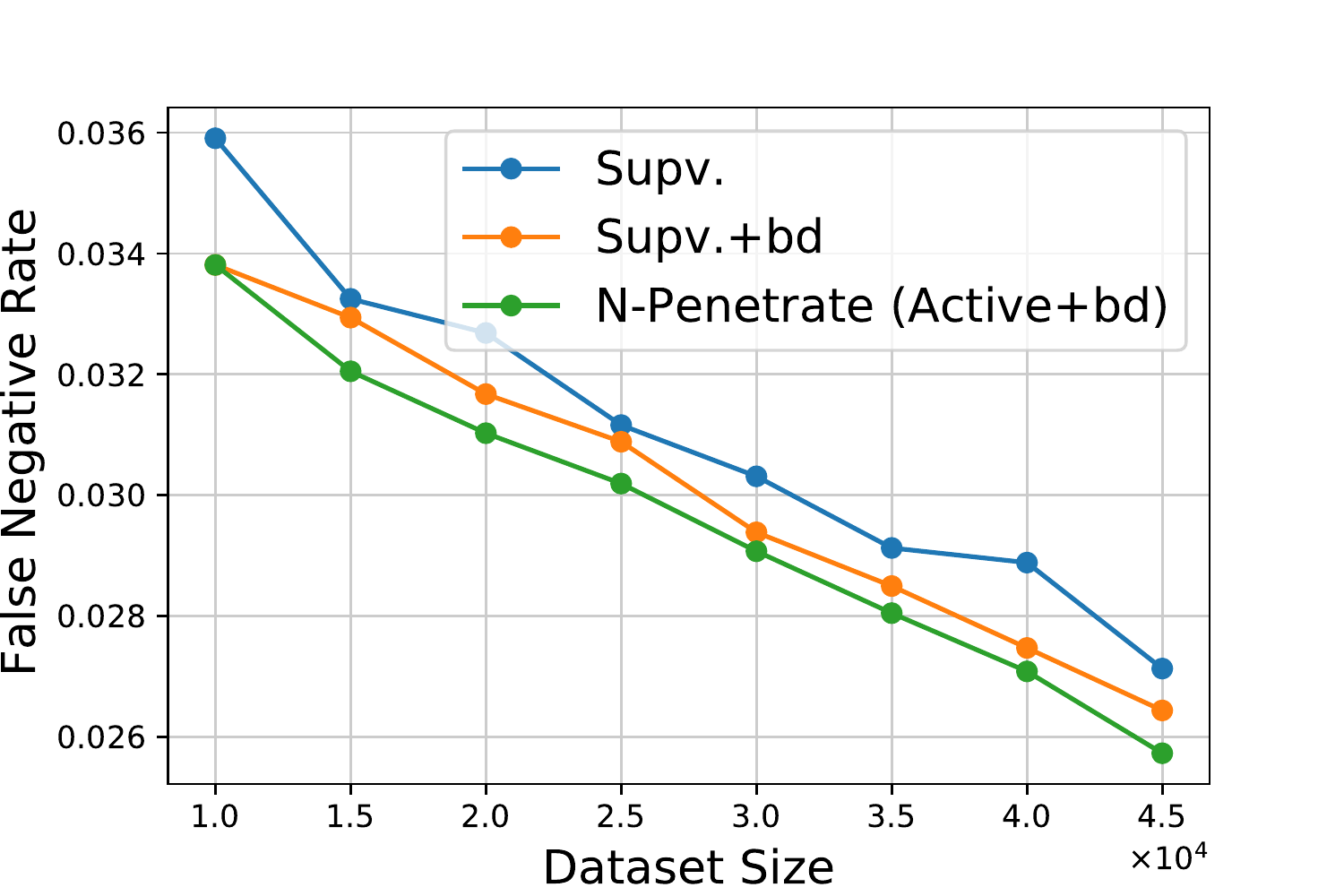}&
\includegraphics[width=0.18\linewidth]{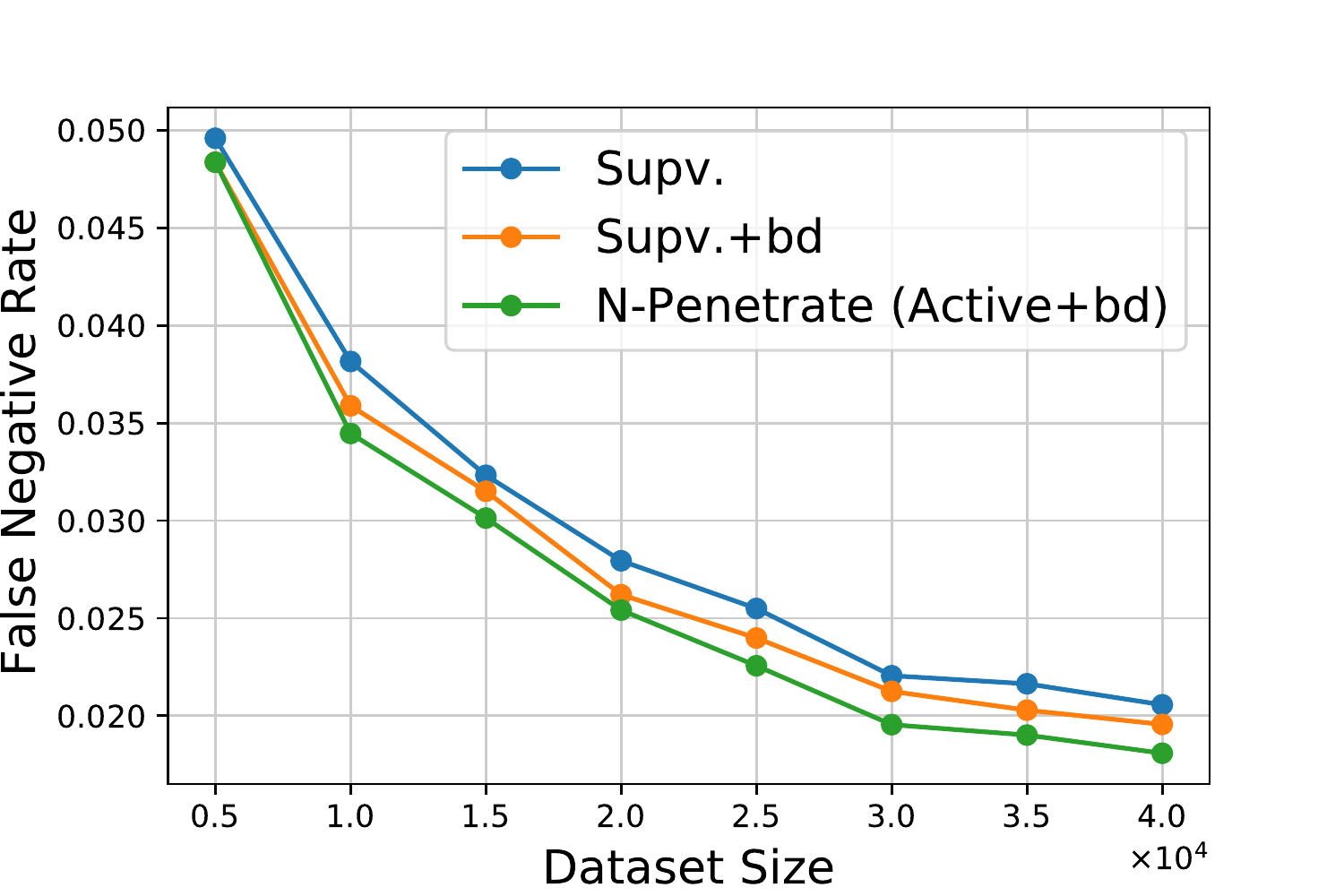}&
\includegraphics[width=0.18\linewidth]{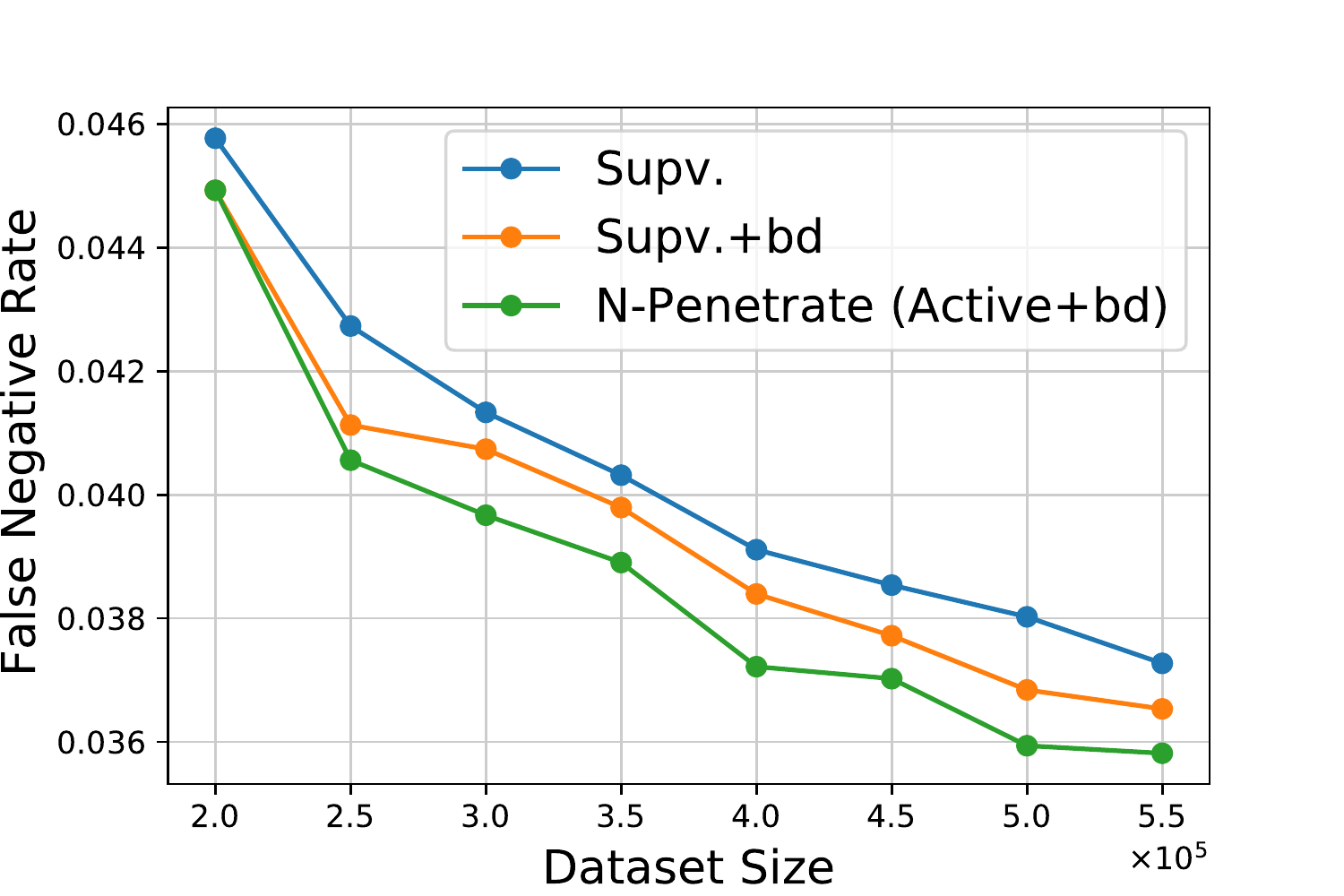}
\end{tabular}
\vspace{-10px}
\caption{\footnotesize{\label{fig:iterationFalseNegativeRate} We plot the false negative rate against the dataset size. The baselines are trained using the same amount of data. On average, ours achieves a $14.27\%$ lower false negative rate than \textbf{\textit{Supv}}. From left to right: SCAPE, Swing, Jump, Skirt, and Hand.}}
\vspace{-10px}
\end{figure*}

\begin{figure*}[h]
\centering
\setlength{\tabcolsep}{2pt}
\begin{tabular}{ccccc}
\includegraphics[width=0.18\linewidth]{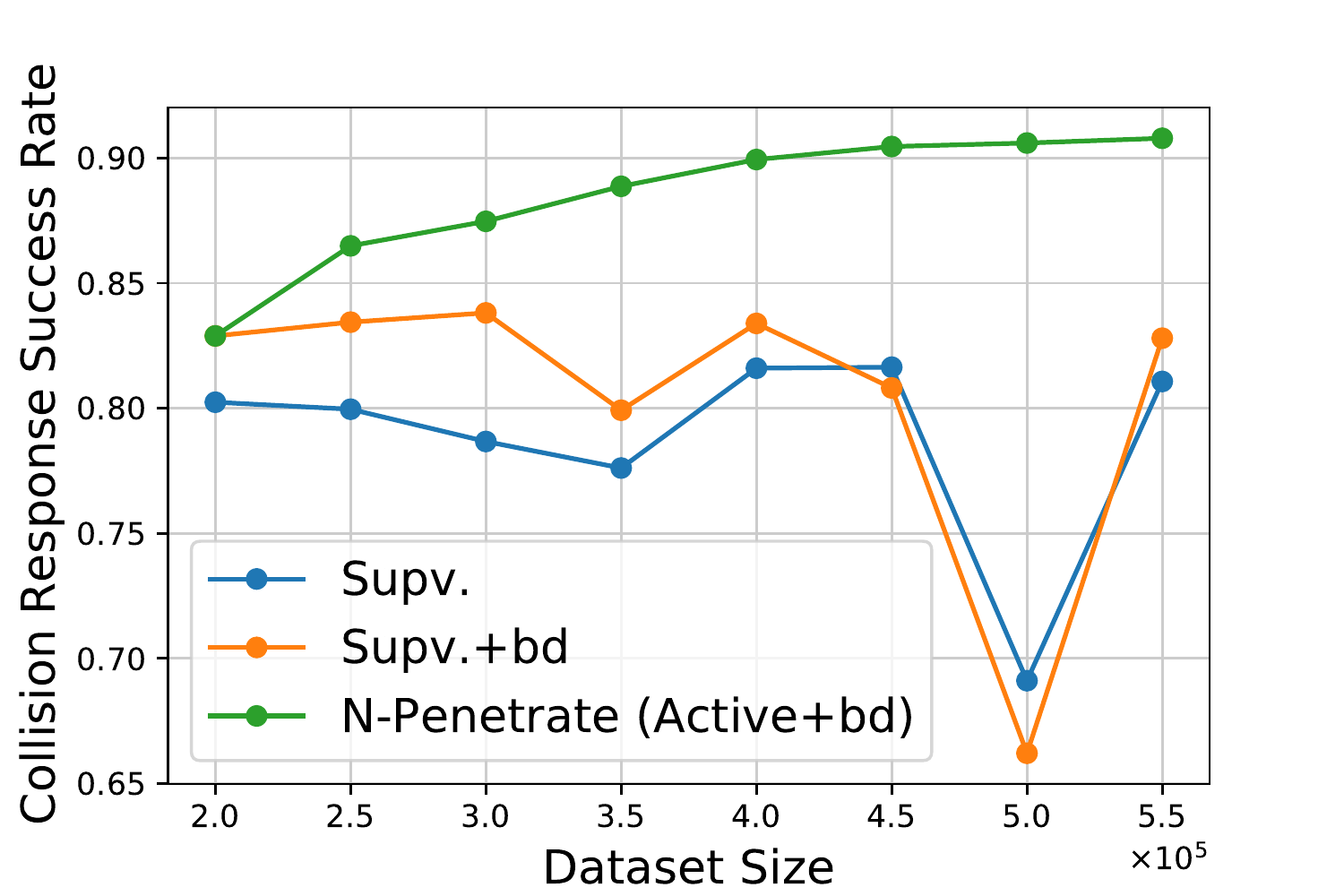}&
\includegraphics[width=0.18\linewidth]{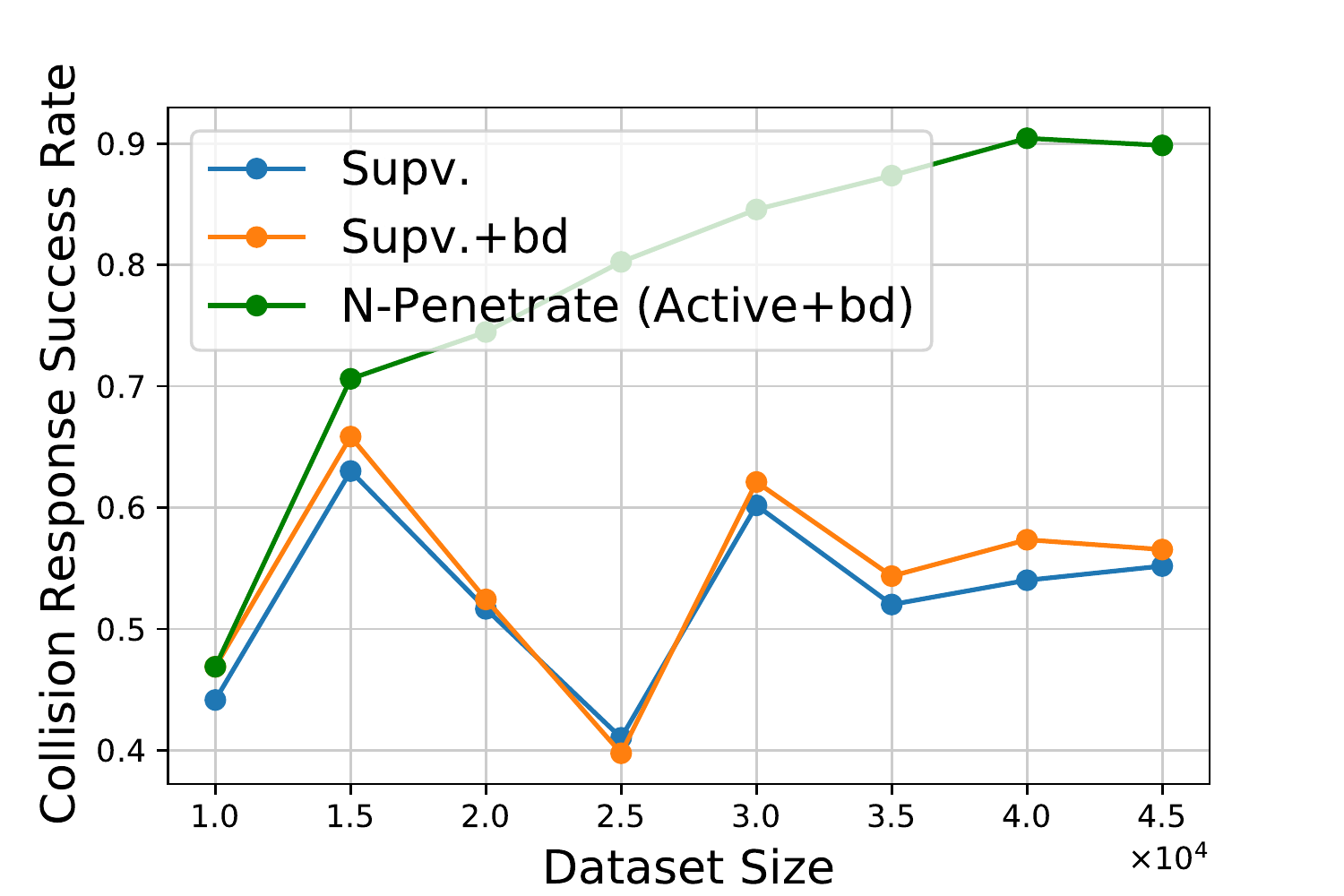}&
\includegraphics[width=0.18\linewidth]{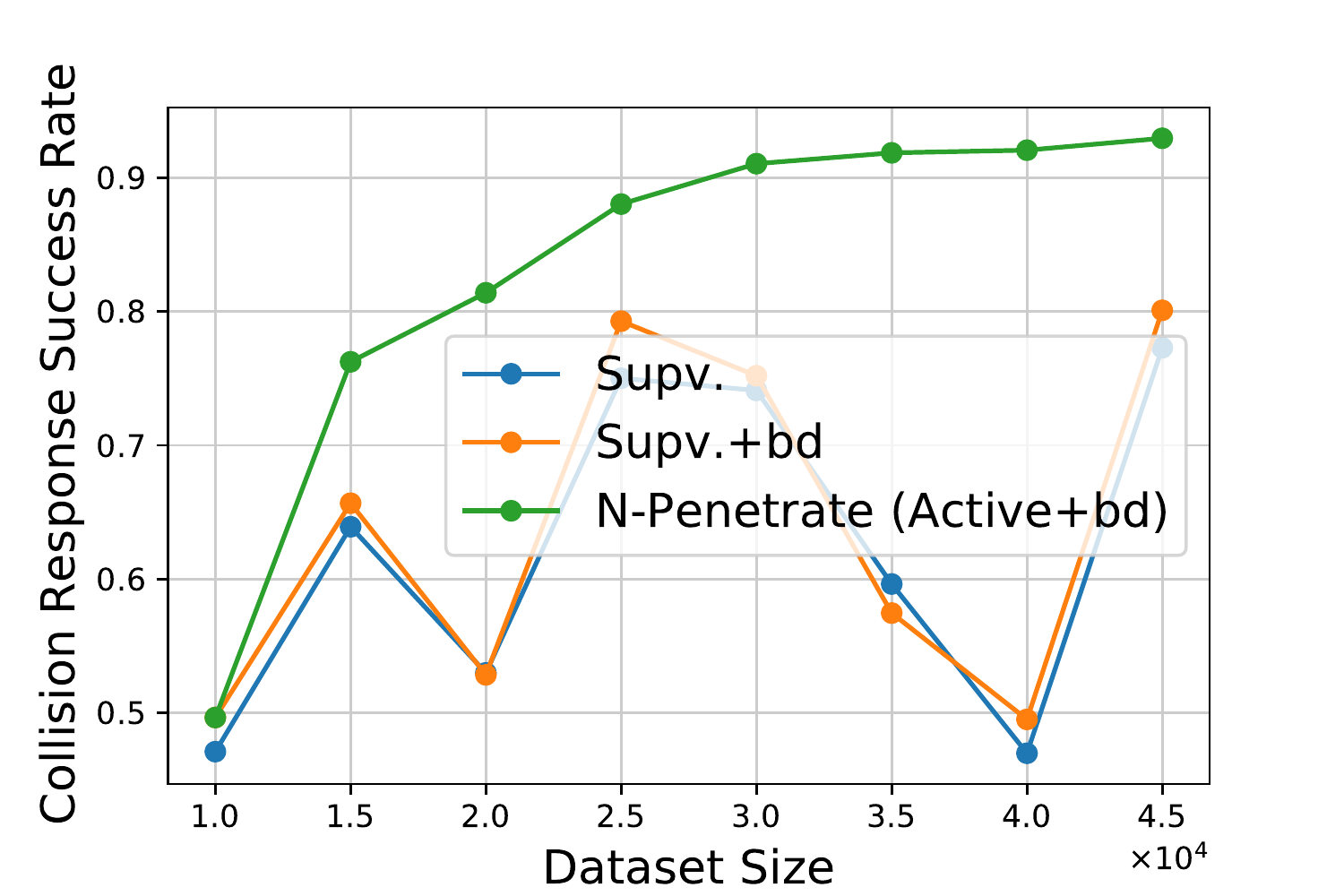}&
\includegraphics[width=0.18\linewidth]{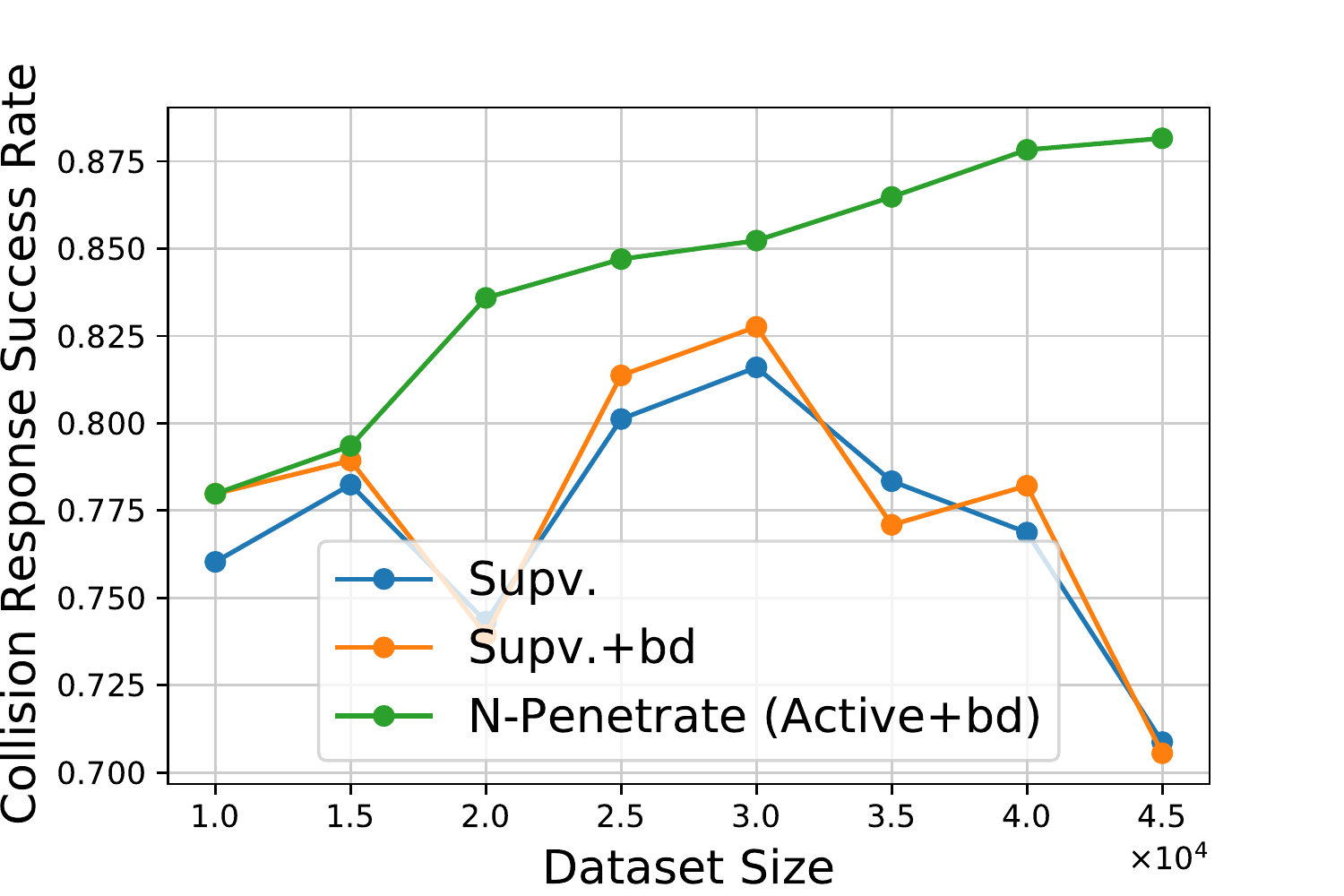}&
\includegraphics[width=0.18\linewidth]{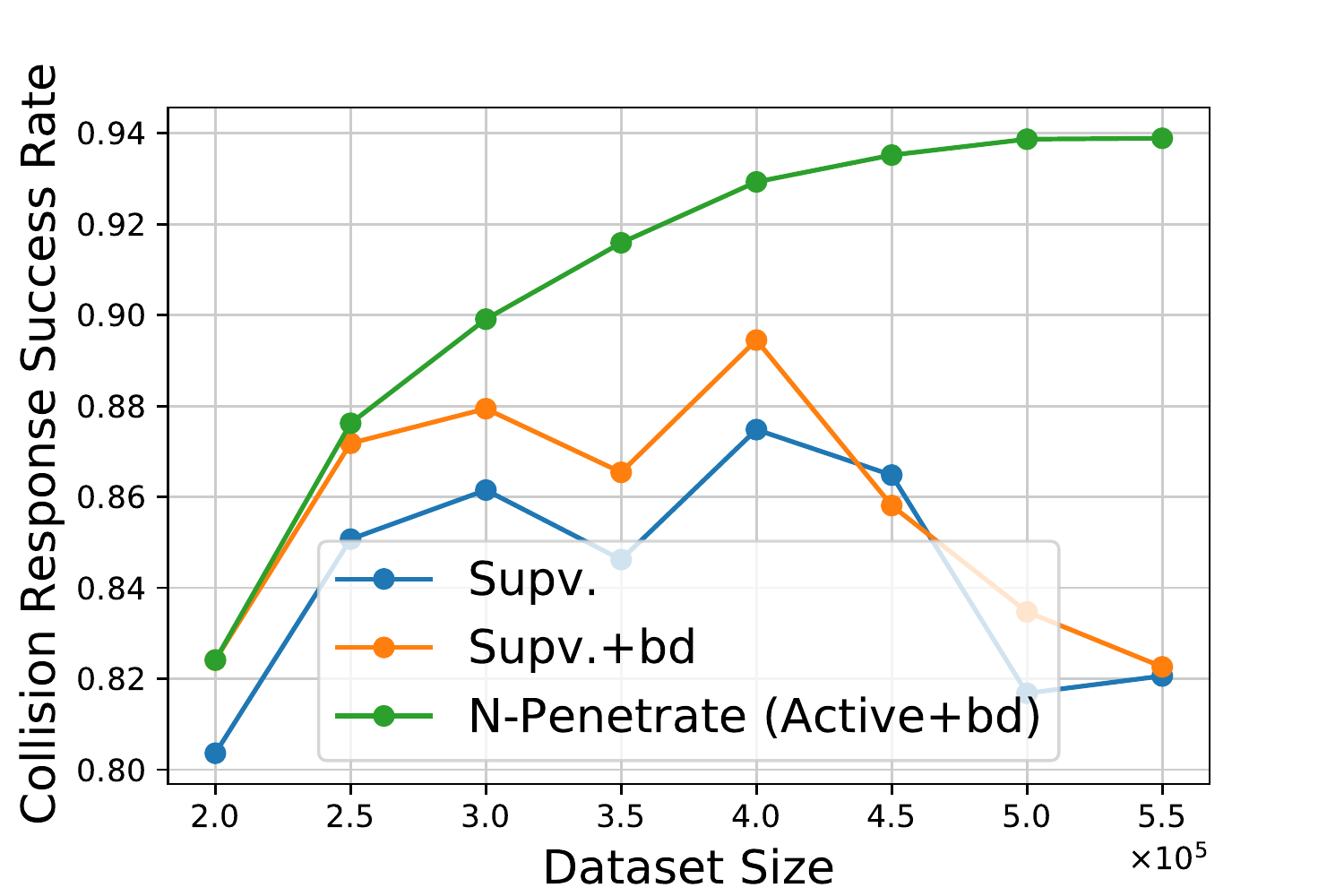}
\end{tabular}
\vspace{-10px}
\caption{\footnotesize{\label{fig:iterationColResSuccRate} We plot the success rate of the neural collision handler against the dataset size. Our method resolves $24.74\%$ more collisions than \textbf{\textit{Supv+bd}}. From left to right: SCAPE, Swing, Jump, Skirt, and Hand.}}
\vspace{-10px}
\end{figure*}

\section{\label{sec:evaluation}Evaluation}
\TE{Datasets:} We evaluate our method on five types of datasets, as illustrated in \prettyref{fig:examples}. The first three (SCAPE \cite{anguelov2005scape} with $N=71$ meshes each having $2161$ vertices, MIT Swing \cite{vlasic2008articulated} with $N=150$ meshes each having $9971$ vertices, and MIT Jump \cite{vlasic2008articulated} with $N=150$ meshes each having $10002$ vertices) contain human bodies with different sets of actions and poses. We have also tested our method on a skirt dataset introduced by \cite{yang2020multiscale} that contains $N=201$ simulated skirt meshes synthesized by NVIDIA clothing tools, each of which has $2830$ vertices. The skirt is deformable everywhere, and the dataset is rather small. Obtaining stable performance in this case is challenging and we observe reasonably good results using active learning. Finally, we introduce a custom dataset of human hand poses. We captured various hand poses and transitions between the poses in a multi-view capture system. We ran 3D reconstruction~\cite{Galliani15} and 3D keypoint detection~\cite{Simon_2017_CVPR} for the captured images and registered a linear blend skinning model consisting of $2825$ vertices for each frame of the data~\cite{gall2009motion}, resulting in $N=7314$ meshes.

\begin{table}[h]
\vspace{-5px}
\centering
\scalebox{0.6}{
\setlength{\tabcolsep}{4pt}
\begin{tabular}{cccccc}
\toprule
&SCAPE & Swing & Jump & Skirt & Hand \\
\midrule
$N_\text{init}$ &200000 & 10000 & 10000 & 5000 & 200000   \\
$N_\text{aug}$ & 50000 & 5000 & 5000 & 5000 & 50000\\
\bottomrule
\end{tabular}}
\vspace{-5px}
\caption{\label{table:Ninit}\small{$N_\text{init}$ and $N_\text{aug}$ used by each dataset.}}
\vspace{-10px}
\end{table}
\TE{Implementation:} We implement our method using PyTorch with the same network architecture as \cite{tan2020lcollision} and perform experiments on a desktop machine with an NVIDIA RTX 2080Ti GPU.  We begin by training $\left<\theta_E,\theta_D\right>$ using Adam with a learning rate of $0.01$ and a batch size of $128$ over $3000$ epochs. For neural collision detector training, unless otherwise stated, we choose the following hyper-parameters: $\epsilon=1\times10^{-4}, w_\text{PD}=5, w_\text{PDsum}=0.2, w_r=2, w_{ce}=2, w_b=0.5$. We perform bootstrap by supervised learning $\theta_C$ on $N_\text{init}$ data points. We initialize $N_\text{init}=2000$ and progressively inject data points into $N_\text{init}$ until the ``elbow point'' of the accuracy vs. the sample size is reached, which is detected using \cite{5961514}. For each experiment, we train $\theta_C$ using Adam with a learning rate of $0.001$ and a batch size of $512$ over $N_\text{epoch}=100$ epochs. We choose suitable $N_\text{aug}$ according to $N_\text{init}$. The $N_\text{init}$ and $N_\text{aug}$ used for each dataset are summarized in \prettyref{table:Ninit}. During data aggregation, we terminate Newton's method when the relative changes of $Z_\text{all}$ are less than $\epsilon_z=10^{-7}$. For each subsequent iteration of active data augmentation, we fine-tune $\theta_C$ using Adam with a learning rate of $10^{-4}$ and a batch size of $1024$ over $N_\text{epoch}=50$ epochs. For collision handling, we run ALM until \prettyref{eq:handler} is satisfied. 

\TE{Collision Detection:} We compare our method with two baseline algorithms. The first one \cite{tan2020lcollision} (denoted as \textbf{\textit{Supv}}) uses the same network architecture as ours with both the autoencoder $\left<\theta_E,\theta_D\right>$ and the collision detector $\theta_C$ trained using supervised learning, where $\mathcal{D}_c$ is constructed by randomly sampled poses from $U(\mathcal{Z})$ and boundary set $D_b$ with associated loss \prettyref{eq:bd_loss} mentioned in \prettyref{sec:bd} is not used, i.e., $w_b=0$. Our second baseline  (denoted as \textbf{\textit{Supv + bd}}) also trains both networks using supervised learning, but the boundary set and loss in \prettyref{eq:bd_loss} are used. Our proposed neural collision handling pipeline also uses active learning and boundary information and is denoted as \textbf{\textit{Active + bd}}. After $k$ iterations of active data augmentations, we have a dataset with $N_{init}+kN_{aug}$ points for training $\theta_C$. For fairness, we re-train our two baselines using $N_{init}+kN_{aug}$ points randomly sampled from $U(\mathcal{Z})$ after each iteration. For all the methods, we use $80\%$ of the data for training and the rest is used as a validation set for hyperparameter tuning. For each dataset, we create a test set with $7.5\times 10^5$ samples from $U(\mathcal{Z})$, which is unseen in the training stage, to evaluate the performances. The performances of neural collision detectors are evaluated based on two metrics: the fraction of successful predicates (accuracy) and the fraction of times a self-penetrating mesh is erroneously predicted as collision-free (false negative rate). False negatives are more detrimental to our applications than false positives as our collision handler only takes care of positive samples. As illustrated in \prettyref{fig:iterationAccuracy} and \prettyref{fig:iterationFalseNegativeRate}, our method effectively improves both metrics. The performance after active learning is summarized in \prettyref{table:perf}. We reach $93.8 - 98.1\%$ accuracy compared to the groundtruth generated by the exact method \cite{pan2012fcl}, with up to 124$\times$ speedup. On average, our method achieves $1.86\%$ higher accuracy and a $14.27\%$ lower false negative rate than \textbf{\textit{Supv}}. In the last row of \prettyref{table:perf}, we measure an equivalent dataset size, which is defined as the size of the dataset needed by the \textbf{\textit{Supv+bd}} to achieve the same accuracy as our method. We derive this number by interpolating on experimental results of the \textbf{\textit{Supv+bd}}. Our method achieves a similar accuracy using a $35.0\%$ smaller dataset than \textbf{\textit{Supv+bd}} on average.
\begin{table}[t]
\centering
\resizebox{0.47\textwidth}{!}{
\setlength{\tabcolsep}{4pt}
\begin{tabular}{cccccc}
\toprule
metric                   & SCAPE & Swing & Jump & Skirt & Hand \\
\midrule
final dataset size       & $5.5\times10^5$ & $4.5\times 10^4$ & $4.5\times 10^4$ & $4\times 10^4$ & $5.5\times10^5$ \\
\midrule
accuracy (Ours (\textbf{\textit{Active+bd}}))           & 0.9383 & 0.9638 & 0.9552 & 0.9817 & 0.9692   \\
accuracy (\textbf{\textit{Supv+bd}})                    & 0.9282 & 0.9609 & 0.9500 & 0.9795 & 0.9650   \\
accuracy (\textbf{\textit{Supv}})                       & 0.9181 & 0.9460 & 0.9347 & 0.9660 & 0.9558   \\
\midrule
false neg. (Ours (\textbf{\textit{Active+bd}}))         & 0.05151 & 0.01485 & 0.02573 & 0.01808 & 0.03582   \\
false neg. (\textbf{\textit{Supv+bd}})                  & 0.05576 & 0.01766 & 0.02644 & 0.01956 & 0.03652   \\
false neg. (\textbf{\textit{Supv}})                     & 0.06914 & 0.01969 & 0.02713 & 0.02056 & 0.03727   \\
\midrule
equi. dataset size (\textbf{\textit{Supv+bd}})          & $6.63\times10^5$ & $7.64\times10^4$ & $7.02\times10^4$ & $7.71\times 10^4$ & $7.30\times10^5$   \\
\bottomrule
\end{tabular}}
\vspace{-5px}
\caption{\label{table:perf}\footnotesize{We summarize the accuracy and false negative rate of three methods under comparison. We also include the equivalent dataset size for the baseline to reach the same performance as our method.}}
\vspace{-10px}
\end{table}

\begin{figure}
\begin{minipage}[b]{0.2\textwidth}
\centering
\includegraphics[width=\textwidth]{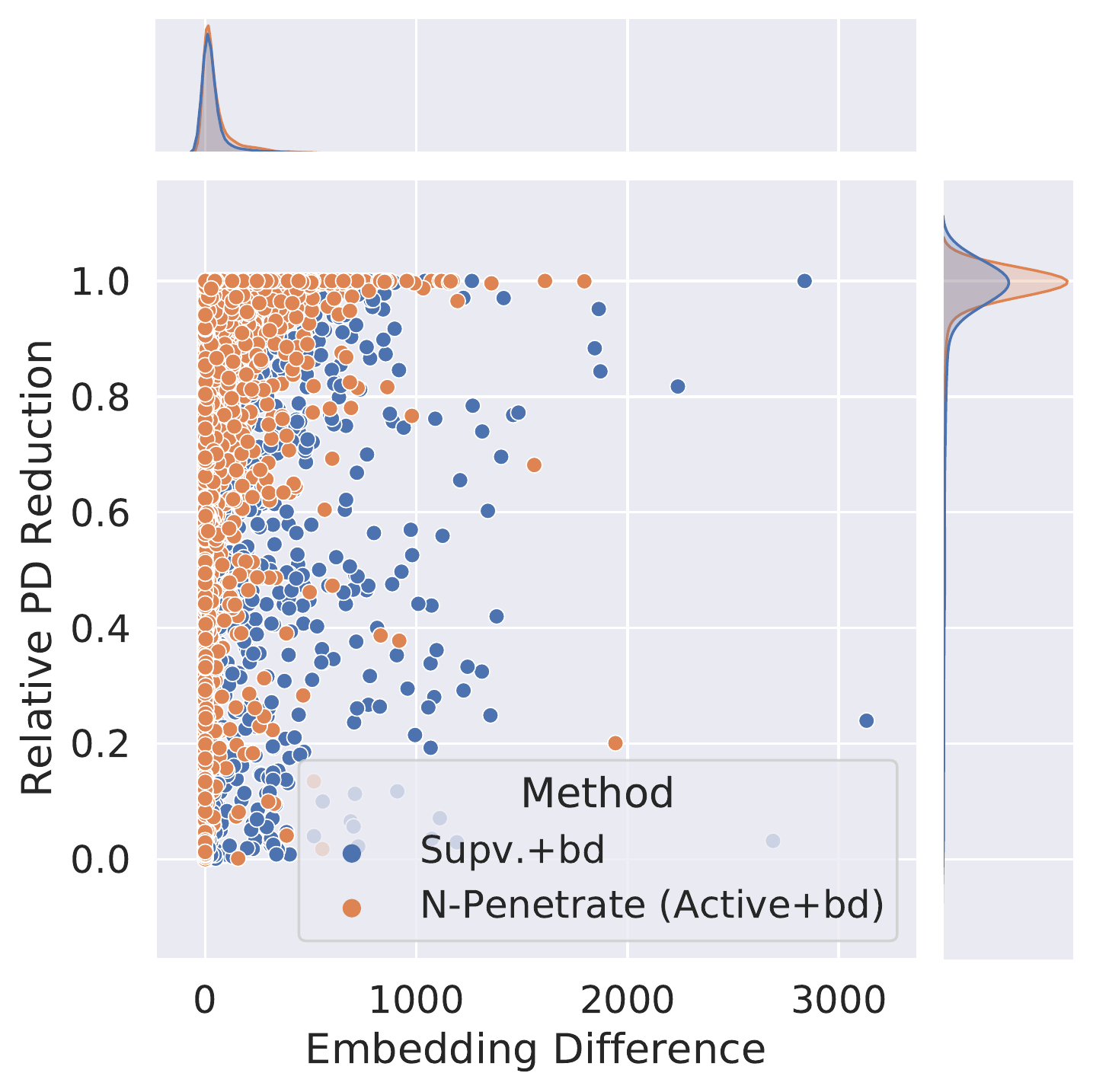}
\vspace{-20px}
\end{minipage}
\hfill
\begin{minipage}[b]{.24\textwidth}
\captionof{figure}{\scriptsize{We plot joint distribution of relative $\text{PD}$ reduction ($Y$-axis) and embedding difference ($X$-axis) over successfully collision-handled test meshes in the SCAPE dataset for our method and \textbf{\textit{Supv+bd}}. Our method resolves more collisions (average PD reduction $94.81\%$ vs. $88.76\%$) while remaining closer to the input (average embedding difference 56.17 vs. 66.11) compared to \textbf{\textit{Supv+bd}}.}}
\label{fig:PDReduction} 
\vspace{-20px}
\end{minipage}
\end{figure}

\TE{Collision Handling:} We plug the trained neural collision detectors into ALM and compare our method, \textbf{\textit{Supv}} and, \textbf{\textit{Supv+bd}} in terms of resolving self-penetrating meshes. To this end, we randomly sample $10000$ self-penetrating, unseen $Z_{all}^{user}$ from $U(\mathcal{Z})$, and use \prettyref{eq:handler} to derive $Z_{all}$. 
We compare the performance based on relative $\text{PD}$ reduction defined as:
\footnotesize
\begin{align*}
\frac{\text{PD}(\text{ACAP}^{-1}(Z_\text{all}^{user},\theta_D))-\text{PD}(\text{ACAP}^{-1}(Z_\text{all},\theta_D))}{\text{PD}(\text{ACAP}^{-1}(Z_\text{all}^{user},\theta_D))}.
\end{align*}
\normalsize
Collision resolution is completely successful if this value equals one, which may not always happen because ALM uses soft penalties to relax hard constraints. Thus, we consider a solution successful if the value is greater than $0$. We plot the success rate against the dataset size in \prettyref{fig:iterationColResSuccRate}, which shows that our method resolves $24.74\%$ more collisions than \textbf{\textit{Supv+bd}}. Thanks to our risk-seeking data aggregation method, our method monotonically improves the collision handling success rate when more data points are injected, while \textbf{\textit{Supv+bd}} exhibits unstable performance. Since \textbf{\textit{Supv}} uses the same randomly sampled dataset as \textbf{\textit{Supv+bd}}, the performance exhibits similar instability. Meanwhile, our novel boundary loss improves the results for \textbf{\textit{Supv+bd}}, since it can better approximate the decision boundary. Another criterion for good collision handling is the embedding difference -- the objective function in \prettyref{eq:handler}. We want the output to be as close as possible to the input. We plot the relative $\text{PD}$ reduction vs. embedding difference over successfully collision-handled test meshes in the SCAPE dataset for our method and \textbf{\textit{Supv+bd}} in \prettyref{fig:PDReduction}. The mean relative $\text{PD}$ reduction for our method is $94.81\%$ and the mean embedding difference is $56.17$, compared to $88.76\%$ and $66.11$, respectively, for \textbf{\textit{Supv+bd}}. The results show that our method resolves more collisions while the outputs stay closer to the input latent codes. Some exemplary results are shown in \prettyref{fig:examples}.
\section{\label{sec:conclusion}Conclusion \& Limitations}
We present an active learning method for training a neural collision detector in which training data are progressively sampled from the learned latent space using a risk-seeking approach. Our approach is designed for general 3D deformable meshes, and we highlight its benefits on many complex datasets. In practice, our method outperforms supervised learning in terms of accuracy, false negative rate, and stability. As a major limitation, our collision handler does not consider physics models. This can be performed in the future by integrating a learning-based physics simulation approach such as \citep{Zheng_2021_CVPR}. We are also considering extensions to meshes with changing topologies, e.g., using a level-set-based mesh representation.

\bibliography{references}
\end{document}